\documentclass[lettersize,journal]{IEEEtran}
\usepackage{amsmath,amsfonts}
\usepackage{algorithmic}
\usepackage{algorithm}
\usepackage{array}
\usepackage[caption=false,font=normalsize,labelfont=sf,textfont=sf]{subfig}
\usepackage{textcomp}
\usepackage{stfloats}
\usepackage{url}
\usepackage{verbatim}
\usepackage{graphicx}
\usepackage{cite}
\def\etal{\textit{et al}.}

\def\eg{\textit{e.g.}}
\usepackage{caption}
\usepackage[table]{xcolor}
\usepackage{multirow}
\usepackage{bm}
\usepackage{dashrule}
\usepackage{booktabs}

\IEEEoverridecommandlockouts

\IEEEaftertitletext{
  \noindent\begin{minipage}{\textwidth}\centering
    \includegraphics[width=\textwidth]{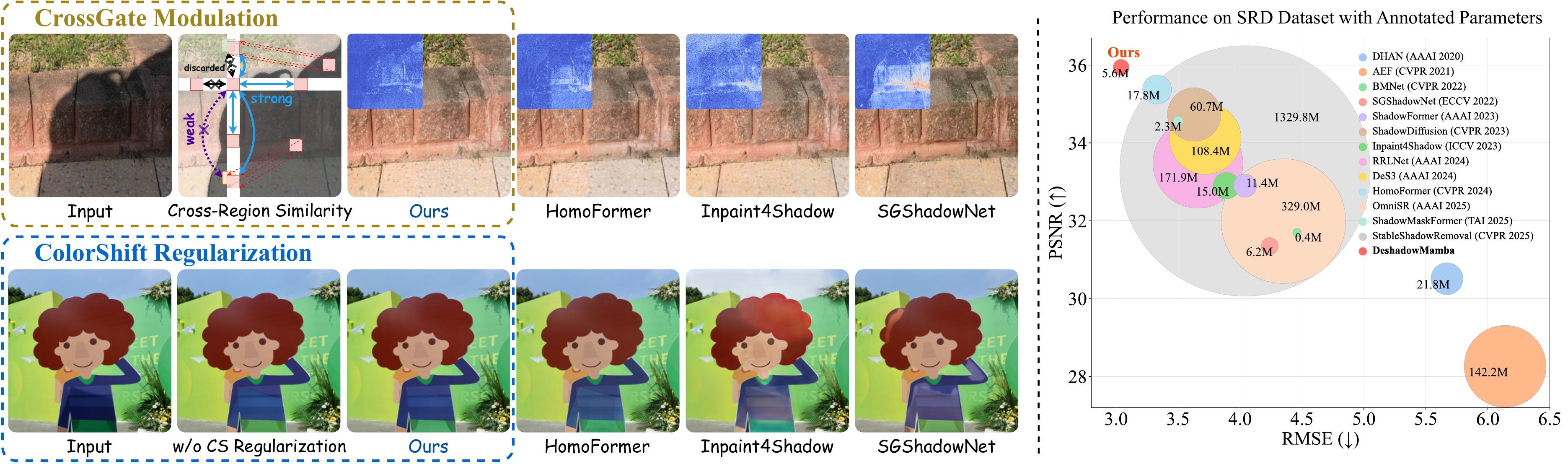}
    \captionsetup{type=figure}
    \captionof{figure}{We revisit image shadow removal from a sequence modeling perspective and present DeshadowMamba, a Mamba-based framework enhanced with CrossGate modulation and ColorShift regularization. Our method produces more accurate structure and color restoration compared to prior approaches, and achieves superior performance with strong parameter efficiency.}
    \label{fig:teaser}
  \end{minipage}
}

\begin{document}

\title{DeshadowMamba: Deshadowing as 1D Sequential Similarity}

\author{Zhaotong Yang, Yi Chen, Yanying Li, Shengfeng He, Yangyang Xu, Junyu Dong, Jian Yang, Yong Du
}

\markboth{}
{Shell \MakeLowercase{\textit{et al.}}: A Sample Article Using IEEEtran.cls for IEEE Journals}

\maketitle

\begin{abstract}
Recent deep models for image shadow removal often rely on attention-based architectures to capture long-range dependencies. However, their fixed attention patterns tend to mix illumination cues from irrelevant regions, leading to distorted structures and inconsistent colors. In this work, we revisit shadow removal from a sequence modeling perspective and explore the use of Mamba, a selective state space model that propagates global context through directional state transitions. These transitions yield an efficient global receptive field while preserving positional continuity. Despite its potential, directly applying Mamba to image data is suboptimal, since it lacks awareness of shadow–non-shadow semantics and remains susceptible to color interference from nearby regions. To address these limitations, we propose CrossGate, a directional modulation mechanism that injects shadow-aware similarity into Mamba’s input gate, allowing selective integration of relevant context along transition axes. To further ensure appearance fidelity, we introduce ColorShift regularization, a contrastive learning objective driven by global color statistics. By synthesizing structured informative negatives, it guides the model to suppress color contamination and achieve robust color restoration. Together, these components adapt sequence modeling to the structural integrity and chromatic consistency required for shadow removal. Extensive experiments on public benchmarks demonstrate that DeshadowMamba achieves state-of-the-art visual quality and strong quantitative performance.
\end{abstract}


\section{Introduction}
\IEEEPARstart{S}{hadows} are common in real-world images and often degrade visual quality while interfering with downstream tasks such as object detection~\cite{gao2024multiscale}, tracking~\cite{li2017weighted}, and appearance manipulation~\cite{lyu2021sogan}. Image shadow removal, which aims to recover a clean image by eliminating shadows, is a fundamental problem in computer vision. The task requires not only restoring occluded content but also maintaining spatial coherence and consistent appearance across shadow boundaries.

Recent learning-based methods~\cite{le2020shadow,fu2021auto,wan2022style,niu2022boundary} have achieved notable progress using convolutional neural networks (CNNs). However, the inherent locality of CNNs limits their ability to model long-range dependencies, which are essential for leveraging non-shadow regions to guide the recovery of shadowed areas. Transformers, with their strong global modeling capability, have become attractive alternatives. Despite their success, practical implementations often rely on window-based or region-limited attention~\cite{liu2024regional} to reduce the quadratic complexity, which fails to provide the full-context awareness needed for localized degradations such as shadows. Some recent works~\cite{xiao2024homoformer} attempt to overcome this limitation through pixel shuffling, which redistributes spatial tokens to encourage cross-region interaction. While such designs extend receptive fields, they inevitably disturb local structure and may lead to spatial misalignment. These trade-offs highlight a key question: \textit{can global context be modeled efficiently while preserving spatial alignment?}

State space models offer a promising direction toward this goal. The S4 model~\cite{gu2021efficiently} and its improved variant, Mamba~\cite{gu2023mamba}, have demonstrated competitive performance in low-level vision tasks~\cite{guo2024mambair,MambaLLIE,dong2024ecmamba,zou2024freqmamba}. In contrast to transformers that rely on window-based attention to reduce complexity, Mamba maintains spatial continuity by propagating information through selective one-dimensional state transitions, thereby attaining a global receptive field with linear complexity. Moreover, this one-dimensional sequential property makes Mamba particularly suitable for shadow removal, where shadows typically exhibit smooth intensity transitions and coherent spatial continuity across regions.

Nevertheless, directly applying Mamba to image shadow removal remains insufficient. Although the limitation of its unidirectional state update can be largely mitigated through multi-directional scanning, the key issue lies in its input gating mechanism, which tends to prioritize high-contrast or salient regions~\cite{han2024demystify} that are not necessarily informative for re-illuminating shadows. In addition, as a model with global feature aggregation, Mamba may entangle color statistics from chromatically irrelevant regions, leading to noticeable color inconsistency in the restored results.

To address these challenges, we present DeshadowMamba, a shadow removal framework that harnesses the strengths of Mamba while fundamentally enhancing its ability to model shadow-specific context. At the core of our design is CrossGate, a directional modulation mechanism that enables deformable modeling of semantic similarities between shadow and non-shadow regions. By injecting these similarity cues into Mamba’s input gate, DeshadowMamba selectively propagates reliable information from non-shadow areas while maintaining stability in degraded regions. This cross-region modulation extends Mamba’s sequential modeling paradigm to spatially structured visual restoration, facilitating stable and context-aware feature refinement in shadow regions. 

We further tackle the persistent issue of color inconsistency between restored shadow regions and their surroundings, which often stems from entangled color statistics in global representations and luminance mismatch caused by incomplete shadow recovery. To mitigate this, we introduce ColorShift regularization, a contrastive learning scheme that constructs informative negative samples through controlled color transformations. By simulating realistic chromatic deviations, ColorShift encourages the network to resist color interference and maintain a coherent appearance across shadow boundaries. Combining CrossGate and ColorShift, DeshadowMamba tailors sequence modeling to both the structural and chromatic demands of image shadow removal, achieving perceptually consistent restoration and competitive quantitative performance across multiple public benchmarks.

The main contributions of this work are summarized as follows:
\begin{itemize}[]
    \item We revisit image shadow removal from a sequence modeling perspective and propose DeshadowMamba, a novel framework grounded in Mamba's state space formulation to achieve efficient global context propagation while explicitly preserving spatial structure.
    \item We propose CrossGate, a directional modulation mechanism that computes shadow-aware similarity between spatial positions and injects it into Mamba’s input gate. This enables selective integration of relevant non-shadow features to guide shadow region reconstruction.
    \item We introduce ColorShift regularization, a contrastive learning strategy driven by global color statistics. By synthesizing structured informative negatives through controlled color shifts, it encourages the model to resist color contamination and improves chromatic consistency.
    \item Extensive experiments on standard benchmarks demonstrate that DeshadowMamba achieves state-of-the-art performance in both visual quality and quantitative metrics.
\end{itemize}

\begin{figure*}[t] 
    \centering
    \includegraphics[width=\textwidth]{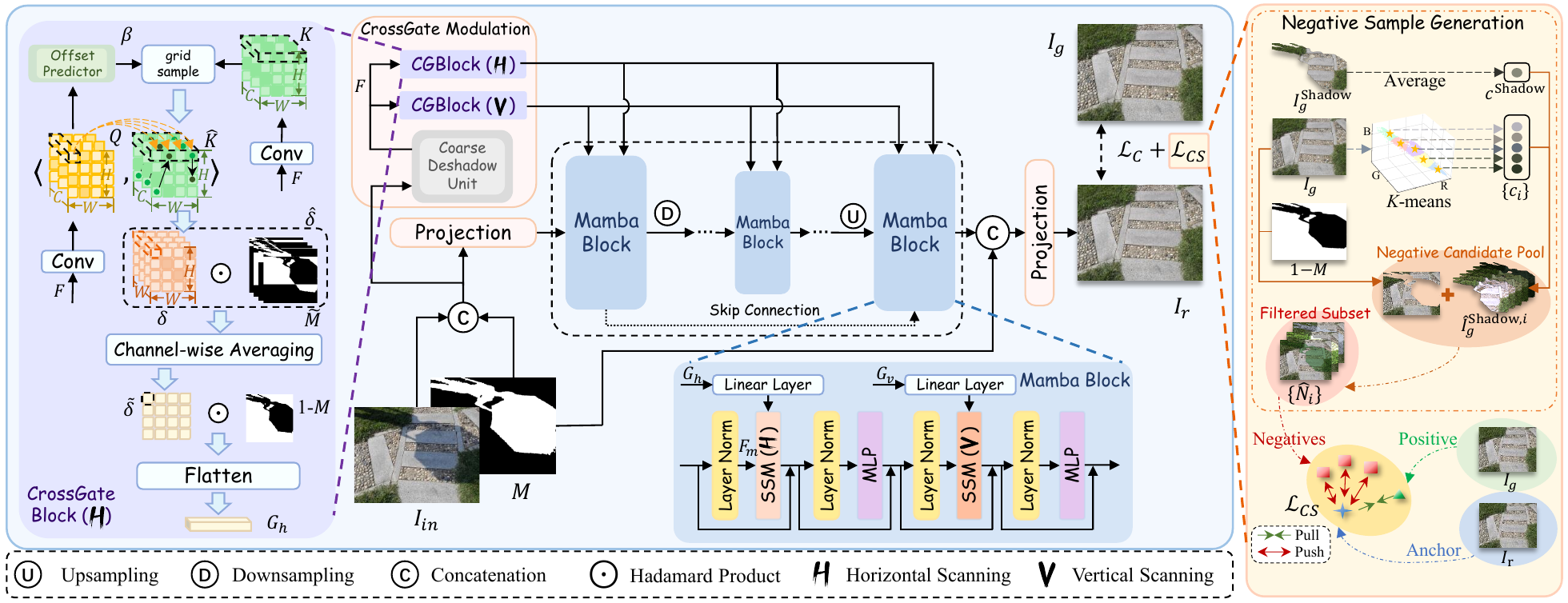}
    \caption{DeshadowMamba consists of a Mamba-based encoder-decoder architecture enhanced by CrossGate modulation and ColorShift regularization. CrossGate injects directional, shadow-aware similarity into Mamba's input gate to guide feature integration, while ColorShift generates weighted contrastive samples to enforce color consistency during training.}\vspace{-4mm}
    \label{fig:network_architecture}
\end{figure*}

\section{Related Work}
\subsection{Image Shadow Removal}  
Image shadow removal is a long-standing task in computer vision. Early traditional methods~\cite{finlayson2009entropy,yang2012shadow,ma2016appearance,xiao2013fast,guo2011single,finlayson2005removal} rely heavily on hand-crafted priors and physical assumptions, but often fail in complex real-world scenes. With the rise of deep learning, shadow removal has made substantial progress. Some methods~\cite{le2020shadow,Zhu_Xiao_Fang_Fu_Xiong_Zha_2022,le2021physics} incorporate physical degradation modeling, while others explore contextual priors such as directional features~\cite{hu2019direction}, exposure fusion~\cite{fu2021auto}, or joint inpainting~\cite{li2023leveraging}. While these priors improve local restoration, they remain limited in capturing broader spatial relationships and maintaining illumination consistency.

To alleviate the reliance on paired supervision, unsupervised generative approaches~\cite{jin2021dc,liu2021shadow} attempt to learn shadow removal directly from unpaired data. Although they reduce annotation costs, their reconstruction quality often lags behind supervised counterparts. Diffusion-based models~\cite{guo2023shadowdiffusion,guo2023boundary} further advance generative shadow removal by introducing degradation priors and boundary constraints, yet their iterative denoising process leads to substantial computational overhead. 

Meanwhile, convolutional networks remain efficient but are inherently limited by their local receptive fields, making it difficult to model global illumination variations. Transformer-based architectures~\cite{guo2023shadowformer,xiao2024homoformer} address this by leveraging self-attention for global context modeling; for instance, Guo \etal~\cite{guo2023shadowformer} enhance channel-wise dependencies, while Xiao \etal~\cite{xiao2024homoformer} introduce pixel shuffling to reduce spatial bias. However, these methods still struggle to jointly balance efficiency, structural preservation, and non-local dependency modeling.

Motivated by these observations, we introduce DeshadowMamba, which revisits shadow removal from a structured sequence modeling perspective. Our method not only preserves spatial continuity but also effectively models long-range dependencies in an efficient and scalable manner, providing a unified solution for shadow removal.

\subsection{State Space Models for Image Restoration}  
Selective state space models such as Mamba~\cite{gu2023mamba} have recently emerged as powerful alternatives to Transformers in vision tasks, offering linear complexity and strong global modeling capabilities. Several works have extended Mamba to low-level image restoration, including super-resolution and denoising~\cite{guo2024mambair}, low-light enhancement~\cite{MambaLLIE}, multi-exposure correction~\cite{dong2024ecmamba}, deblurring~\cite{gao2024learning}, and deraining~\cite{zou2024freqmamba}. While these methods demonstrate promising results on global degradations, they primarily focus on full-image enhancement and are less effective for localized degradations such as shadows, which require fine-grained spatial and chromatic corrections. In this work, we extend Mamba’s directional modeling capability with task-specific components that enable shadow-aware context modulation and localized appearance correction.

\subsection{Contrastive Learning}  
Contrastive learning has gained popularity in low-level vision tasks~\cite{zheng2023curricular,gao2024efficient,hang2022scs,liang2022semantically,zhang2024unified,huang2023contrastive} due to its ability to learn discriminative representations by contrasting positive and negative samples. A key challenge, however, lies in constructing meaningful contrastive pairs that reflect realistic degradation variations. Liang \etal~\cite{liang2022semantically} simulate exposure variations to improve robustness in low-light enhancement, while Hang \etal~\cite{hang2022scs} synthesize feature-level negatives with inconsistent styles to enhance contrastive discrimination. Zheng \etal~\cite{zheng2023curricular} aggregate predictions from multiple dehazing models to establish a consensus-based contrastive space. Despite these advances, contrastive learning remains underexplored in image shadow removal. We address this gap through ColorShift regularization, which constructs structured and informative negative samples via controlled color perturbations in shadow regions. This formulation encourages the model to distinguish valid color cues from corrupted ones, achieving consistent and faithful color restoration across shadow boundaries.

\section{Preliminaries}
In sequence modeling, state-space models (SSMs) establish a mathematical framework that unifies continuous-time dynamics with discrete-time sequence processing. Given a one-dimensional input signal $x(t)$, the system maintains a hidden state $\bm{h}(t)\in\mathbb{R}^Z$ and produces an output $\bm{y}(t)$, governed by the continuous-time equations:  
\begin{equation}
\begin{aligned}
& \bm{h}'(t) = \bm{Ah}(t) + \bm{B}x(t), \
& \bm{y}(t) = \bm{Ch}(t) + Dx(t),
\end{aligned}
\end{equation}
where $\bm{A}\in\mathbb{R}^{Z\times Z}$ defines state transitions, $\bm{B}\in\mathbb{R}^{Z\times 1}$ encodes input dynamics, $\bm{C}\in\mathbb{R}^{1\times Z}$ maps hidden states to outputs, and $D\in\mathbb{R}$ represents direct feedthrough.  

To adapt SSMs to the discrete-time setting used in deep learning, the zero-order hold (ZOH) method is commonly employed for discretization. Given a time step $\bm{\Delta}$, the discrete parameters are computed as:  
\begin{equation}
\overline{\bm{A}} = \exp(\bm{\Delta A}), \quad
\overline{\bm{B}} = (\bm{\Delta A})^{-1}(\exp(\bm{\Delta A}) - \mathbb{I}) \bm{\Delta B}.
\end{equation}
This leads to the following discrete-time recursion:  
\begin{equation}
\begin{aligned}
& \bm{h}_t = \overline{\bm{A}}\bm{h}_{t-1} + \overline{\bm{B}}x_t, \
& \bm{y}_t = \bm{Ch}_t + Dx_t.
\end{aligned}
\end{equation}

In conventional discrete SSMs, parameters $\{\overline{\bm{A}}, \overline{\bm{B}}, \bm{C}, D\}$ remain fixed, resulting in time-invariant processing. Mamba~\cite{gu2023mamba} breaks this constraint through input-dependent parameterization, which is defined as follows: 
\begin{equation}
\begin{aligned}
& \bm{B} = S_{\bm{B}}(x), \quad \bm{C} = S_{\bm{C}}(x), \
& \bm{\Delta} = \mathrm{Softplus}\left( \theta_{\bm{\Delta}} + S_{\bm{\Delta}}(x) \right),
\end{aligned}
\end{equation}
where $\theta_{\bm{\Delta}}$ is a learnable bias term and $S_{\cdot}(\cdot)$ denotes linear projection layers. This selective mechanism enables context-dependent state transitions and adaptive memory dynamics across time. More recently, state-space duality (SSD)~\cite{dao2024transformers} has further accelerated Mamba by reformulating temporal recursions into parallelizable matrix operations, allowing efficient GPU execution without compromising modeling capacity.

\section{Method}
\subsection{Overview}
Given a shadow image $\bm{I}_{in}$ and a shadow mask $\bm{M}$, DeshadowMamba aims to produce a shadow-free result $\bm{I}_r$ by leveraging Mamba's sequence modeling capability. We introduce a \textit{CrossGate} modulation strategy that injects point-wise sequential similarity into Mamba while preserving its global receptive field. It enables the model to capture long-range dependencies along both horizontal and vertical directions through gated one-dimensional interactions. 

To address color shifting artifacts, including hue deviations induced by interference from non-shadow regions and luminance inconsistencies caused by imperfect shadow recovery, we introduce a \textit{ColorShift} regularization guided by global color statistics. This mechanism mitigates shadow–color contamination and enhances the fidelity of appearance restoration.

As illustrated in Fig.~\ref{fig:network_architecture}, the overall architecture consists of a stack of Mamba blocks, augmented with a lightweight direction-aware enhancement module, referred to as the CrossGate design, which jointly performs coarse shadow removal and contextual modulation. The ColorShift regularization is applied during training to reinforce intra-region fidelity and reduce boundary inconsistencies. Together, these components form a cohesive framework for effective and perceptually robust shadow removal. Notably, the shadow mask $\bm{M}$ can be obtained from manual annotations or automatic shadow detectors~\cite{vicente2016large}, making the method applicable to real-world scenarios. 

\subsection{CrossGate Modulation} 
In the Mamba formulation, the step size $\bm{\Delta}$ controls the balance between the current input and the historical state~\cite{gu2023mamba}. A larger $\bm{\Delta}$ increases the contribution of the input signal to the state update, whereas a smaller one enhances the influence of the previous hidden state. Modulating $\bm{\Delta}$ with cues from non-shadow regions is therefore crucial for leveraging reliable inputs while avoiding degraded ones in shadowed areas. To achieve this, we propose CrossGate, a modulation module that adapts Mamba’s input gate using shadow-aware non-local similarity. It consists of a coarse deshadow unit, composed of two Mamba blocks that efficiently provide shadow-suppressed features for reliable similarity estimation, and two directional CrossGate blocks for horizontal and vertical scanning. 

We take the horizontal CrossGate block as an example, while its vertical counterpart performs a column-wise operation following the same principle. Given a feature map $\bm{F}\!\in\!\mathbb{R}^{C\times H\times W}$ extracted from the coarse deshadow unit, two independent convolutional layers $l_q(\cdot)$ and $l_k(\cdot)$ are employed to generate the query and key feature embeddings:
\begin{equation}
  \bm{Q}=l_q(\bm{F}), \quad \bm{K}=l_k(\bm{F}).
\end{equation}
Using separate layers allows the query and key to learn complementary representations, which enables more discriminative similarity estimation for identifying task-relevant correlations. 

We compute point-wise similarity between $\bm{Q}$ and $\bm{K}$ to capture cross-region correlations that provide cues from non-shadow areas. However, directly evaluating all spatial pairs would incur quadratic computational complexity, which contradicts the efficiency advantage of Mamba. To alleviate this, we draw inspiration from empirical studies~\cite{guo2024mambair}, which reveal that Mamba’s strongest activations are concentrated within a cross-shaped region corresponding to its four directional scanning paths (top-left~$\leftrightarrow$~bottom-right and their rotated counterparts). This observation implies that Mamba’s state transitions inherently emphasize horizontally and vertically aligned dependencies. Leveraging this property, CrossGate restricts similarity computation to the same row in the horizontal block and to the same column in the vertical counterpart, preserving Mamba’s directional inductive bias while maintaining efficiency.

Nevertheless, a fixed row-wise sampling pattern may fail to capture semantically correlated regions beyond the current row, making it difficult to assess the reliability of similarity-based cues. To address this, we introduce a lightweight offset predictor $O(\cdot)$ that estimates learnable offsets $\bm{\beta}=O(\bm{Q})$, adaptively warping feature responses onto the row-aligned sampling field. These offsets expand the effective receptive field of similarity estimation, enabling CrossGate to capture non-local dependencies in a direction-aware and scalable manner. The offsets are applied to the key features through deformable sampling:
\begin{equation}
    \bm{\hat{K}}=\operatorname{grid\_sample}(\bm{K},\bm{\beta}).
\end{equation} 
For each spatial position $(i,j)$, we compute the similarity between its query feature $\bm{Q}_{:,i,j}$ and all warped key features $\hat{\bm{K}}_{:,i,r}$ along the same row-aligned direction:
\begin{equation}
\bm{\delta}_{r,i,j} = \langle \bm{Q}_{:,i,j}, \hat{\bm{K}}_{:,i,r} \rangle,
\label{eq:5}
\end{equation}
where $\langle \cdot, \cdot \rangle$ denotes the inner product, $i\!\in\!\{1,\dots,H\}$ and $j,r\!\in\!\{1,\dots,W\}$ index the spatial positions of the query and warped key features, respectively. The resulting tensor $\bm{\delta}\!\in\!\mathbb{R}^{W\times H\times W}$ encodes the row-wise correlations between each query location and all deformed key positions.

To isolate cross-region similarities that provide meaningful restoration cues, we construct a binary gating tensor $\tilde{\bm{M}}\!\in\!\mathbb{R}^{W\times H\times W}$ from the input shadow mask $\bm{M}\!\in\!\mathbb{R}^{H\times W}$. Specifically, only correlations between shadow and non-shadow regions are preserved, while intra-region similarities are masked out, as they contribute little to restoration and may propagate degraded information. To ensure spatial alignment between the gating and the deformed similarity field, the learned offsets are applied to $\bm{M}$, yielding a warped mask $\hat{\bm{M}}$ corresponding to $\bm{\hat{K}}$. The cross-region indicator tensor is then defined as
\begin{equation}
\tilde{\bm{M}}_{r,i,j}=\bm{M}_{i,j} \oplus \hat{\bm{M}}_{i,r},\quad
\hat{\bm{\delta}} = \bm{\delta} \odot \tilde{\bm{M}}, 
\label{eq:6}
\end{equation}
where $\oplus$ and $\odot$ denote XOR and Hadamard operations, respectively. The filtered similarity tensor $\hat{\bm{\delta}}$ therefore retains only cross-region correlations between shadow and non-shadow areas, ensuring that the modulation focuses exclusively on task-relevant long-range dependencies.

To obtain a compact modulation signal compatible with the input-gate computation, we aggregate the filtered similarity tensor by performing channel-wise averaging:
\begin{equation} 
\tilde{\bm{\delta}}_{i,j} = \frac{1}{W} \sum_{r=1}^{W} \hat{\bm{\delta}}_{r,i,j}. 
\label{eq:7} 
\end{equation}
This operation summarizes the shadow-aware similarity responses into a direction-aware relevance map $\tilde{\bm{\delta}}\!\in\!\mathbb{R}^{H\times W}$, indicating the strength of non-local cues at each spatial position. Since larger values correspond to greater reliance on the input signal, we retain only the responses from non-shadow pixels to obtain the horizontal modulation signal $\bm{G}_h$:
\begin{equation} 
\bm{G}_h = \tilde{\bm{\delta}} \odot (1 - \bm{M}).
\label{eq:8} 
\end{equation}
This selective modulation enables the gating process to leverage clean-region cues while maintaining stability in degraded regions.

Finally, $\bm{G}_h$ is linearly projected and injected into the input-gate computation of the horizontal SSD module in the Mamba block:
\begin{equation} 
\bm{\Delta} = \operatorname{Softplus}\left(\theta_{\bm{\Delta}} + S_{\bm{\Delta}}(\bm{F}_m) + S_{\bm{G}}(\bm{G}_h)\right), 
\label{eq:9} 
\end{equation}
where $\bm{F}_m$ denotes the input features of the SSD module, and $S_{\bm{\Delta}}(\cdot)$ and $S_{\bm{G}}(\cdot)$ are linear layers. The same procedure is applied in the vertical direction, yielding $\bm{G}_v$ to complete the CrossGate modulation. 

Note that CrossGate requires only two modulation signals, $\bm{G}_h$ and $\bm{G}_v$, to regulate the step size across Mamba’s four scanning paths. This compact design leverages CrossGate’s ability, absent in standard Mamba, to capture both past and future dependencies within each direction through cross-region similarity estimation, making additional modulation branches unnecessary. More importantly, CrossGate enables Mamba to differentiate the relevance of non-shadow regions during shadow recovery, effectively suppressing less informative local responses and prioritizing semantically meaningful yet spatially distant cues, all while maintaining its global modeling capability.

\subsection{ColorShift Regularization}
Current shadow removal methods often suffer from boundary inconsistencies, primarily because the restored shadow regions fail to reproduce faithful colors. We propose ColorShift Regularization, which leverages contrastive learning to alleviate this issue. A key challenge lies in constructing effective contrastive samples. Specifically, we use the ground-truth image as a positive sample and generate structured negative samples by applying controlled color shifts. These shifts simulate potential chromatic deviations and define a bounded deshadow solution space for anchor traversal.

For each ground-truth image $\bm{I}_g$ in the training set, we first use the classic $K$-Means~\cite{macqueen1967some} algorithm in the RGB space to extract the dominant color components, yielding $K$ color triplets $\{c_i\}_{i=1}^K$. Additionally, we compute the average color $c^\text{Shadow}$ for $\bm{I}_g^\text{Shadow}$, the shadow region of $\bm{I}_g$:
\begin{equation}
		\bm{I}_g^{\text{Shadow}} = \bm{I}_g\odot \bm{M}, 	      
       c^{\text{Shadow}} = \frac{1}{|\bm{I}_g^{\text{Shadow}}|} \sum \bm{I}_g^{\text{Shadow}},
		\label{eq:9}
\end{equation}
where $|\cdot|$ denotes the total number of pixels. 

Next, we perform a controlled color transformation on $\bm{I}_g^{\text{Shadow}}$, shifting the overall color $c^{\text{Shadow}}$ toward the dominant colors to simulate incorrect color interference. For each color $c_i$, we compute the ratio $r_i\!=\!c_i/c^{\text{Shadow}}$, which serves as a chromatic scaling factor bridging $c_i$ and $c^{\text{Shadow}}$. Taking advantage of the linear properties of the RGB space, we adjust the shadow region by scaling each pixel value with the factor $r_i$: 
\begin{equation} 
\hat{\bm{I}}_g^{\text{Shadow},i} = \bm{I}_g^{\text{Shadow}}\cdot r_i.
\end{equation} 
Consequently, the negative samples $\{\bm{N}_i\}_{i=1}^K$ are synthesized by blending the color-adjusted shadow region $\hat{\bm{I}}_g^{\text{Shadow},i}$ with the non-shadow area of $\bm{I}_g$ for each dominant color $c_i$:
\begin{equation}
	\bm{N}_i=\hat{\bm{I}}_{g}^{\text{Shadow},i}+\bm{I}_{g}\odot(1-\bm{M}).
	\label{eq:10}
\end{equation}
Note that to prevent overflow, pixel values in $\hat{\bm{I}}_{g}^{\text{Shadow},i}$ are clamped to the range $[0,255]$. 

Although these synthesized variants introduce color distortions, their utility for contrastive learning is inconsistent. Some samples exhibit exaggerated shifts that are trivially separable from the ground truth and thus provide limited supervision, whereas others are nearly indistinguishable and overly challenging. To obtain a more meaningful training set, we apply a difficulty-aware filtering strategy to the candidate pool $\{\bm{N}_i\}$. Specifically, we evaluate the difference between each $\bm{N}_i$ and the ground truth $\bm{I}_g$ using the Root Mean Square Error (RMSE) in the LAB color space, denoted as $R_i$. This metric serves as a proxy for learning difficulty. We then preserve only those with $R_i$ falling within the interval 
$(R_{\mu} - R_{\sigma}, R_{\mu} + R_{\sigma})$, where $R_{\mu}$ and $R_{\sigma}$ are the mean and standard deviation of $\{R_i\}_{i=1}^K$, respectively. This filtering step effectively removes outliers and yields a subset $\hat{\{\bm{N_i}\}}$ with balanced difficulty and informative color shifts.

While this filtering improves the overall utility of negative samples, it also leads to a varying number of retained negatives across different shadow inputs. Assigning uniform weights to all negatives under such imbalance can lead to fluctuating gradient magnitudes and degrade training stability. To mitigate this, we assign each filtered negative sample $\hat{\bm{N}}_i$ a weight $\gamma_i$ according to its learning difficulty. Specifically, we normalize the reciprocal of $R_i$ as:
\begin{equation}
\gamma_i = \frac{1/R_i}{\sum_{j=1}^T 1/R_j},
\label{eq:13}
\end{equation}
where $T$ is the total number of valid negatives. This weighting scheme emphasizes moderately hard examples, enabling the network to focus on informative contrasts while avoiding overfitting to easy cases.

Finally, our ColorShift regularization is defined as follows:
\begin{equation}
		\mathcal{L}_{CS}=\frac{\|\bm{f}-\bm{f}^{+}\|_{1}}{\|\bm{f}-\bm{f}^{+}\|_{1}+\sum_{i=1}^{T}\gamma_i\|\bm{f}-\bm{f}_i^{-}\|_{1}},
		\label{eq:14}
\end{equation}
where $\bm{f}=V(\bm{I}_r\odot \bm{M})$ denotes the anchor feature, $\bm{f}^{+}=V(\bm{I}_g\odot \bm{M})$ indicates the positive feature, $\bm{f}_i^{-}=V(\hat{\bm{N}_i}\odot \bm{M})$ corresponds to the negative features. Here, $V(\cdot)$ represents the pre-trained VGG-16~\cite{2015Very} feature extractor, utilizing the output from the $10$th layer. Note that we disable this regularization for patches without any shadow pixels.

\subsection{Training Strategy}
The training of DeshadowMamba proceeds in two stages. In the first stage, to prevent the CrossGate modulation from being influenced by degraded shadow features, we first train a simple single-layer encoder-decoder architecture for coarse shadow removal. This phase employs a Charbonnier loss~\cite{charbonnier1994two}, $\mathcal{L}_C$, to constrain the reconstruction fidelity, defined as: 
\begin{equation} 
\mathcal{L}_C=\sqrt{\|\bm{I}_r-\bm{I}_g\|^2+\epsilon^2}, 
\label{eq:15} 
\end{equation} 
where $\epsilon$ is a small positive constant, set to $10^{-3}$, to ensure computational stability.

In the second stage, we freeze the coarse shadow remover and focus on training the remaining components of the framework, integrating ColorShift regularization. The total learning objective $\mathcal{L}$ for this stage is defined as: 
\begin{equation} 
\mathcal{L}=\mathcal{L}_C + \lambda\mathcal{L}_{CS}, 
\label{eq:16} 
\end{equation} 
where $\lambda$ is a hyperparameter that balances the contributions of the two loss terms.
\section{Experiments}  

\begin{table*}[ht]
    \centering
    \caption{Quantitative evaluations with state-of-the-art methods on the SRD dataset. The best and second-best results are \textbf{bold} and \underline{underlined}, respectively. ``-'' indicates the official source code is not available.}
    \label{tab:exp_srd}
    \resizebox{\textwidth}{!}{
        \begin{tabular}{l|c|ccc|ccc|ccc|c|c}
            \toprule
            \multirow{2}{*}{Method} & \multirow{2}{*}{Venue}
            & \multicolumn{3}{c|}{Shadow} 
            & \multicolumn{3}{c|}{Non-Shadow} 
            & \multicolumn{3}{c|}{All} 
            & \multirow{2}{*}{Params(M)}
            & \multirow{2}{*}{MACs(G)}\\
            \cmidrule(lr){3-5} \cmidrule(lr){6-8} \cmidrule(lr){9-11}
            & & RMSE$\downarrow$ & PSNR$\uparrow$ & SSIM$\uparrow$
            & RMSE$\downarrow$ & PSNR$\uparrow$ & SSIM$\uparrow$
            & RMSE$\downarrow$ & PSNR$\uparrow$ & SSIM$\uparrow$ 
            & & \\
            \midrule
            DHAN~\cite{cun2020towards} & AAAI 2020        & 8.94 & 33.67 & 0.978  & 4.80 & 34.79 & 0.979  & 5.67 & 30.51 & 0.949  & 21.8 & 29.92\\
            \rowcolor[HTML]{f2f2f2}
            AEF~\cite{fu2021auto} & CVPR 2021   & 7.97 & 32.05 & 0.955  & 5.30 & 31.75 & 0.939  & 6.14 & 28.26 & 0.866 & 142.2 & 83.06\\
            BMNet~\cite{zhu2022bijective} & CVPR 2022  & 6.61 & 35.05 & 0.981  & 3.61 & 36.02 & 0.982  & 4.46 & 31.69 & 0.956 & 0.4 & 14.57\\
            \rowcolor[HTML]{f2f2f2}
            SGShadowNet~\cite{wan2022style} & ECCV 2022    & 6.52 & 33.44 & 0.968  & 3.14 & 37.18 & 0.982  & 4.24 & 31.35 & 0.934 & 6.2 & 13.29\\  
            ShadowFormer~\cite{guo2023shadowformer} & AAAI 2023      & 5.90 & 36.91 & 0.982  & 3.44 & 36.22 & 0.983  & 4.04 & 32.90 & 0.957  & 11.4 & 21.05\\
            \rowcolor[HTML]{f2f2f2}
            ShadowDiffusion~\cite{guo2023shadowdiffusion} & CVPR 2023 & 4.98 & 38.72 & \textbf{0.987}  & 3.44 & 37.78 & 0.985  & 3.63 & 34.73 & 0.970 & 60.7 & 182.47\\
            Inpaint4Shadow~\cite{li2023leveraging} & ICCV 2023 & 5.39 & 35.70 & 0.974  & 3.14 & 37.40 & 0.983  & 3.89 & 32.90 & 0.943  & 15.0 & 81.18\\
            \rowcolor[HTML]{f2f2f2}
            RRLNet~\cite{liu2024recasting} & AAAI 2024   & 5.49 & 36.51 & 0.983  & 3.00 & 37.71 & 0.986  & 3.66 & 33.48 & 0.967  & 171.9 & -\\
            DeS3~\cite{jin2024des3} & AAAI 2024   & 5.88 & 37.45 & 0.984  & \underline{2.83} & 38.12 & \underline{0.988}  & 3.72 & 34.11 & 0.968  & 108.4 & 290.53\\
            \rowcolor[HTML]{f2f2f2}
            HomoFormer~\cite{xiao2024homoformer} & CVPR 2024 & \underline{4.25} & \underline{38.81} & \textbf{0.987}  & 2.85 & \underline{39.45} & \underline{0.988}  & \underline{3.33} & \underline{35.37} & \underline{0.972} & 17.8 & 11.93\\
            OmniSR~\cite{xu2024omnisr} & AAAI 2025 & 6.11 & 34.58 & 0.972 & 3.47 & 36.85 & 0.982 & 4.35 & 31.99 & 0.941 & 329.0 & 129.27 \\
            \rowcolor[HTML]{f2f2f2}
            ShadowMaskFormer~\cite{li2025shadowmaskformer} & IEEE TAI 2025 & 4.83 & 37.42 & 0.980 & 2.88 & 39.14 & 0.986 & 3.50 & 34.56 & 0.958 & 2.28 & 12.47 \\
            StableShadowRemoval~\cite{xu_2025_CVPR} & CVPR 2025 & 5.19 & 36.38 & 0.974 & 3.42 & 37.58 & 0.983 & 4.04 & 33.28 & 0.945 & 1329.8 & 30.67\\
            \midrule
            \rowcolor[HTML]{f2f2f2}
            Ours     &       & \textbf{4.09} & \textbf{39.17} & \underline{0.986}  & \textbf{2.52} & \textbf{40.37} & \textbf{0.993}  & \textbf{3.04} & \textbf{35.94} & \textbf{0.974} & 5.6 & 13.52\\
            \bottomrule
        \end{tabular}
    }
\end{table*}

\subsection{Experimental Settings}
\subsubsection{Implementation Details} We implement DeshadowMamba in PyTorch 2.3.1 and train it on a workstation equipped with two NVIDIA GeForce RTX 3090 GPUs. The AdamW~\cite{loshchilov2018decoupled} optimizer is adopted with an initial learning rate of $4\times10^{-4}$, which is decayed to $10^{-6}$ using a cosine annealing schedule~\cite{loshchilov2017sgdr}. The hyperparameter $\lambda$ and the number of color clusters $K$ in the ColorShift regularization are empirically set to $0.01$ and $10$, respectively. Further architecture details and training configurations are provided in the supplementary material.

\subsubsection{Datasets}We evaluate DeshadowMamba on two standard benchmark datasets for single-image shadow removal. The first dataset is SRD~\cite{qu2017deshadownet}, which contains 2,680 training pairs and 408 testing pairs, each comprising a shadow image and its corresponding shadow-free ground truth. Since ground-truth shadow masks are unavailable in SRD, we follow common practice~\cite{xiao2024homoformer,guo2023shadowformer,li2023leveraging,fu2021auto} and utilize masks predicted by DHAN~\cite{cun2020towards} during both training and testing. The second dataset, ISTD+~\cite{le2019shadow}, extends the original ISTD~\cite{wang2018stacked} by correcting illumination inconsistencies between shadow and non-shadow regions. It includes 1,870 paired samples, with 1,330 for training and 540 for testing. Additionally, we evaluate our model on the real-world SBU dataset~\cite{vicente2016large} to assess its robustness and generalization capability under diverse illumination and scene conditions.

\subsubsection{Evaluation Metrics} To ensure a fair comparison, all predicted shadow-free images and ground-truth counterparts are resized to $256\times256$ for evaluation, following prior works~\cite{xiao2024homoformer,guo2023shadowformer,guo2023shadowdiffusion,wan2022style}. The root-mean-square error (RMSE) in the LAB color space serves as the primary evaluation metric. In addition, we also report the peak signal-to-noise ratio (PSNR) and structural similarity index (SSIM) in the RGB color space, consistent with previous studies~\cite{wan2022style,li2023leveraging,liu2024regional,xiao2024homoformer}. For real-world datasets without ground truths, we further employ two no-reference perceptual quality metrics, NIQE and NIMA~\cite{talebi2018nima}, to evaluate perceptual realism and aesthetic consistency. Unless otherwise specified, the results of competing methods are directly cited from their official publications or reproduced using their released code.
\subsection{Comparison with State-of-the-Art Methods}
\begin{figure*}[htbp]
  \captionsetup{skip=2pt}
  \centering
  \newlength{\gutter}
  \setlength{\gutter}{0.0055\textwidth}
  \newlength{\colw}
  \setlength{\colw}{\dimexpr(\textwidth - 7\gutter)/8\relax}

  \newcommand{\imgcell}[1]{%
    \begin{minipage}{\colw}\centering
      \includegraphics[width=\linewidth]{#1}%
    \end{minipage}%
  }
    \newlength{\capht}
    \setlength{\capht}{2.2\baselineskip}
    
    \newcommand{\imgcellcap}[2]{%
      \begin{minipage}[t]{\colw}\centering
        \includegraphics[width=\linewidth]{#1}\par
        \vspace{0.2ex}%
        {\scriptsize
          \parbox[c][\capht][c]{\linewidth}{\centering #2}%
        }%
      \end{minipage}%
    }

  \imgcell{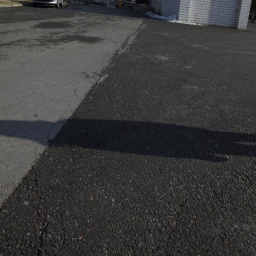}\hspace{\gutter}%
  \imgcell{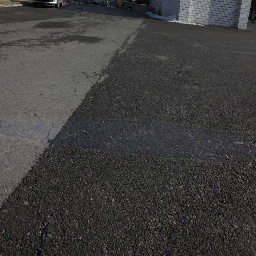}\hspace{\gutter}%
  \imgcell{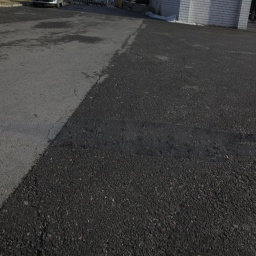}\hspace{\gutter}%
  \imgcell{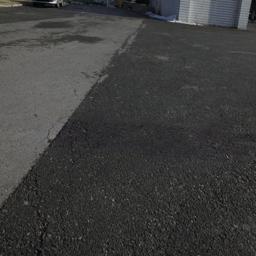}\hspace{\gutter}%
  \imgcell{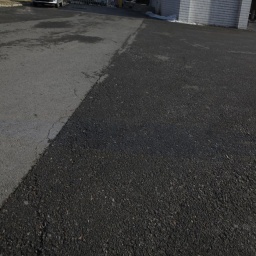}\hspace{\gutter}%
  \imgcell{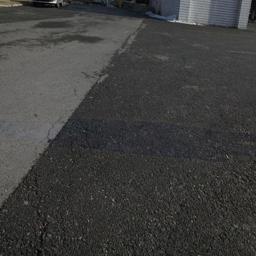}\hspace{\gutter}%
  \imgcell{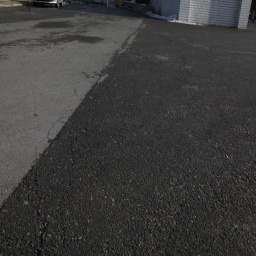}\hspace{\gutter}%
  \imgcell{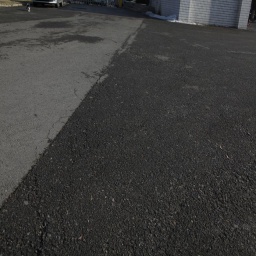}%

  \vspace{0.6em}

  \imgcellcap{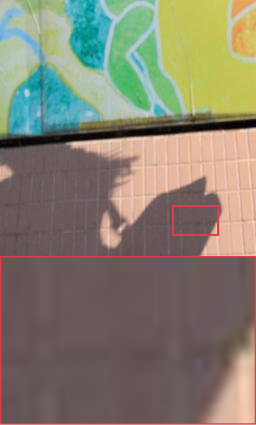}{(a) Shadow Image}\hspace{\gutter}%
  \imgcellcap{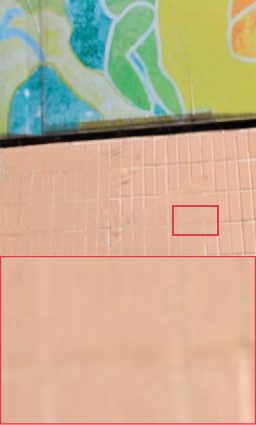}{(b) AEF~\cite{fu2021auto}}\hspace{\gutter}%
  \imgcellcap{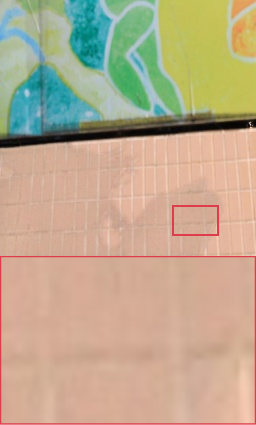}{(c) SGShadowNet~\cite{wan2022style}}\hspace{\gutter}%
  \imgcellcap{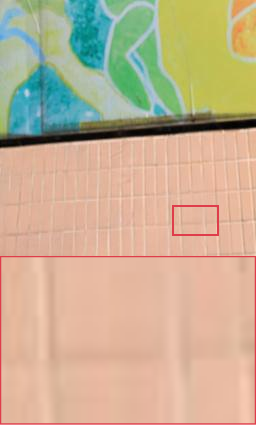}{\shortstack[c]{(d) Inpaint4-\\Shadow~\cite{li2023leveraging}}}\hspace{\gutter}%
  \imgcellcap{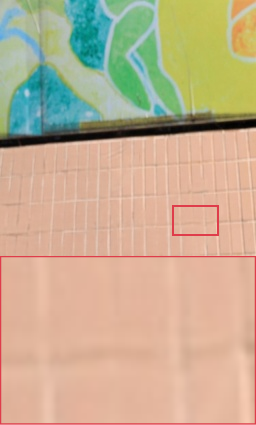}{(e) HomoFormer~\cite{xiao2024homoformer}}\hspace{\gutter}%
  \imgcellcap{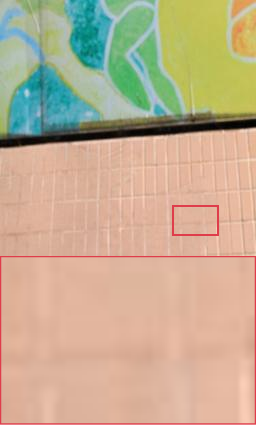}{\shortstack[c]{(f) ShadowMask\\Former~\cite{li2025shadowmaskformer}}}\hspace{\gutter}%
  \imgcellcap{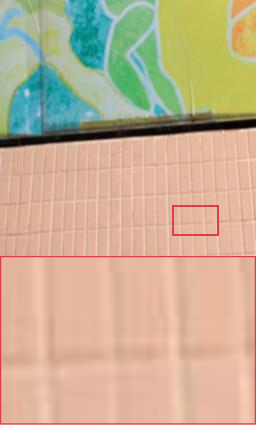}{(g) Ours}\hspace{\gutter}%
  \imgcellcap{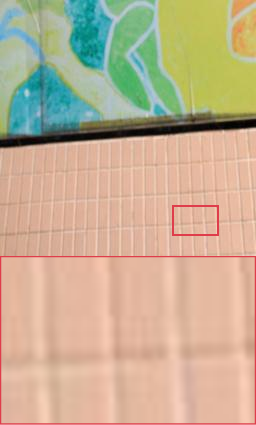}{(h) Ground Truth}%

  \caption{Visual comparisons with state-of-the-art methods on the SRD dataset. (Best viewed zoomed in.)}
  \label{fig:exp_srd}
\end{figure*}

\subsubsection{Evaluation on the SRD Dataset} We compare our method with a wide range of state-of-the-art approaches on the SRD dataset, with quantitative results summarized in Tab.~\ref{tab:exp_srd}. As shown, DeshadowMamba consistently outperforms all competitors, achieving the best RMSE and PSNR values in both shadowed and non-shadowed regions, as well as for the overall image quality. Specifically, for the overall image, our method improves RMSE by $0.29$ and PSNR by $0.57$ dB compared to the second-best method, HomoFormer~\cite{xiao2024homoformer}, clearly demonstrating its superior capability in shadow removal.

For qualitative comparison, Fig.~\ref{fig:exp_srd} shows visual results on the SRD dataset. DeshadowMamba restores fine-grained textures and preserves boundary consistency, producing visually coherent and realistic shadow-free results with minimal color discrepancy across shadow transitions. Beyond accuracy, our model remains compact, with only $5.6$M parameters and comparable MACs, yet maintains state-of-the-art results. Such efficiency and restoration quality together demonstrate the practical advantages of our design.

\begin{table*}[ht]
    \centering
    \caption{Quantitative evaluation with state-of-the-art methods on the ISTD+ dataset.}
    \label{tab:exp_istd+}
    \resizebox{\textwidth}{!}{
        \begin{tabular}{l|c|ccc|ccc|ccc|c|c}
            \toprule
            \multirow{2}{*}{Method} & \multirow{2}{*}{Venue}
            & \multicolumn{3}{c|}{Shadow} 
            & \multicolumn{3}{c|}{Non-Shadow} 
            & \multicolumn{3}{c|}{All} 
            & \multirow{2}{*}{Params(M)}
            & \multirow{2}{*}{MACs(G)}\\
            \cmidrule(lr){3-5} \cmidrule(lr){6-8} \cmidrule(lr){9-11}
            & & RMSE$\downarrow$ & PSNR$\uparrow$ & SSIM$\uparrow$
            & RMSE$\downarrow$ & PSNR$\uparrow$ & SSIM$\uparrow$
            & RMSE$\downarrow$ & PSNR$\uparrow$ & SSIM$\uparrow$ 
            & & \\
            \midrule
            DHAN~\cite{cun2020towards} & AAAI 2020       & 9.60 & 32.98 & 0.987  & 7.44 & 27.12 & 0.973  & 7.80 & 25.65 & 0.955 & 21.8 & 29.92\\
            \rowcolor[HTML]{f2f2f2}
            AEF~\cite{fu2021auto} & CVPR 2021    & 6.64 & 36.18 & 0.977  & 3.75 & 31.31 & 0.884  & 4.20 & 29.59 & 0.849 & 142.2 & 83.06\\
            BMNet~\cite{zhu2022bijective} & CVPR 2022  & 6.24 & 37.37 & 0.990  & 2.46 & 37.85 & 0.984  & 3.02 & 33.95 & 0.967 & 0.4 & 14.57\\
            \rowcolor[HTML]{f2f2f2}
            SGShadowNet~\cite{wan2022style} & ECCV 2022    & 6.46 & 36.91 & 0.989  & 2.95 & 35.47 & 0.976  & 3.45 & 32.46 & 0.956 & 6.2 & 13.29\\
            ShadowFormer~\cite{guo2023shadowformer} & AAAI 2023 & 5.34 & 39.54 & \underline{0.992}  & 2.34 & 38.72 & 0.984  & 2.81 & 35.44 & 0.972 & 11.4 & 21.05\\
            \rowcolor[HTML]{f2f2f2}
            ShadowDiffusion~\cite{guo2023shadowdiffusion} & CVPR 2023 & 4.97 & 39.69 & \underline{0.992}  & 2.28 & 38.89 & \textbf{0.987}  & 2.72 & \underline{35.67} & \textbf{0.975} & 60.7 & 182.47\\
            Inpaint4Shadow~\cite{li2023leveraging} & ICCV 2023 & 6.12 & 38.09 & 0.989  & 2.92 & 36.95 & 0.977  & 3.43 & 33.81 & 0.960 & 15.0 & 81.18\\
            \rowcolor[HTML]{f2f2f2}
            RRLNet~\cite{liu2024recasting}  & AAAI 2024  & 5.69 & 38.04 & 0.990  & 2.31 & \underline{39.15} & 0.984  & 2.87 & 34.96 & 0.968 & 171.9 & -\\
            DeS3~\cite{jin2024des3} & AAAI 2024   & 6.57 & 36.38 & 0.988  & 3.45 & 34.00 & 0.966  & 3.98 & 30.97 & 0.946 & 108.4 & 290.53\\
            \rowcolor[HTML]{f2f2f2}
            HomoFormer~\cite{xiao2024homoformer} & CVPR 2024 & 4.92 & 39.51 & 0.991  & 2.27 & 38.65 & 0.982  & \underline{2.68} & 35.32 & 0.970 & 17.8 & 11.93\\
            OmniSR~\cite{xu2024omnisr} & AAAI 2025 & 6.55 & 37.07 & 0.992 & 2.44 & 37.72 & 0.982 & 3.12 & 33.34 & 0.968 & 329.0 & 129.27 \\
            \rowcolor[HTML]{f2f2f2}
            ShadowMaskFormer~\cite{li2025shadowmaskformer} & IEEE TAI 2025 & 5.46 & 38.79 & 0.991 & \textbf{2.25} & 38.82 & 0.984 & 2.76 & 35.03 & 0.970 & 2.28 & 12.47 \\
            StableShadowRemoval~\cite{xu_2025_CVPR} & CVPR 2025 & \underline{4.43} & \underline{40.02} & \textbf{0.993} & 2.67 & 37.93 & 0.982 & 2.94 & 35.16 & \underline{0.971} & 1329.8 & 30.67\\
            \midrule
            \rowcolor[HTML]{f2f2f2}
            Ours     &      & \textbf{4.35} & \textbf{40.82} & \textbf{0.993}  & \textbf{2.18} & \textbf{39.16} & \underline{0.985}  & \textbf{2.53} & \textbf{36.14} & \textbf{0.975} & 5.6 & 13.52\\
            \bottomrule
        \end{tabular}
    }
\end{table*}

\begin{figure*}[htbp]
  \captionsetup{skip=2pt}
  \centering
  \setlength{\gutter}{0.0055\textwidth}
  \setlength{\colw}{\dimexpr(\textwidth - 7\gutter)/8\relax}

  \newcommand{\imgcell}[1]{%
    \begin{minipage}{\colw}\centering
      \includegraphics[width=\linewidth]{#1}%
    \end{minipage}%
  }
   
   \setlength{\capht}{2.2\baselineskip}
    
    \newcommand{\imgcellcap}[2]{%
      \begin{minipage}[t]{\colw}\centering
        \includegraphics[width=\linewidth]{#1}\par
        \vspace{0.2ex}%
        {\scriptsize
          \parbox[c][\capht][c]{\linewidth}{\centering #2}%
        }%
      \end{minipage}%
    }

  \imgcell{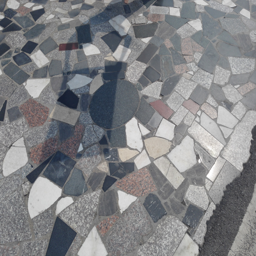}\hspace{\gutter}%
  \imgcell{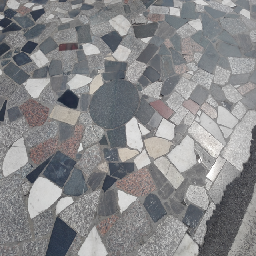}\hspace{\gutter}%
  \imgcell{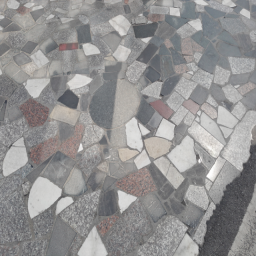}\hspace{\gutter}%
  \imgcell{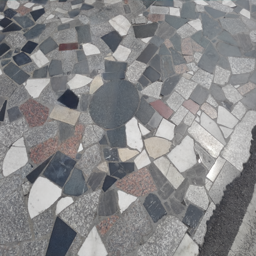}\hspace{\gutter}%
  \imgcell{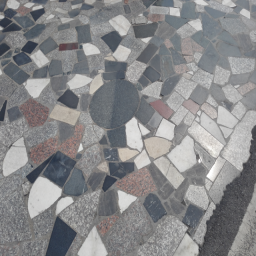}\hspace{\gutter}%
  \imgcell{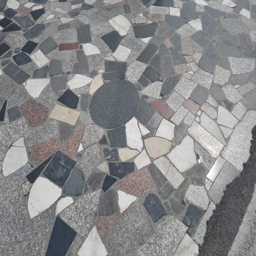}\hspace{\gutter}%
  \imgcell{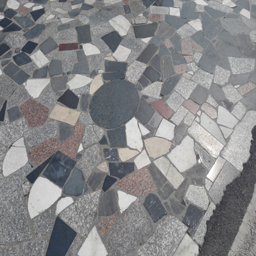}\hspace{\gutter}%
  \imgcell{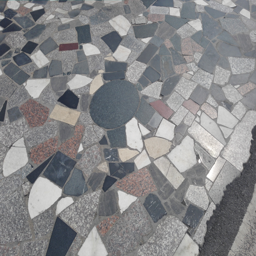}%

  \vspace{0.6em}

  \imgcellcap{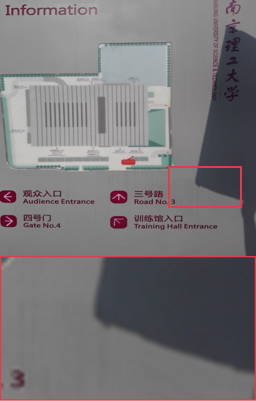}{(a) Shadow Image}\hspace{\gutter}%
  \imgcellcap{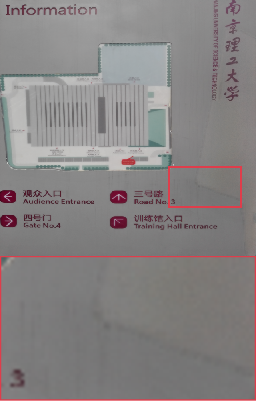}{(b) AEF~\cite{fu2021auto}}\hspace{\gutter}%
  \imgcellcap{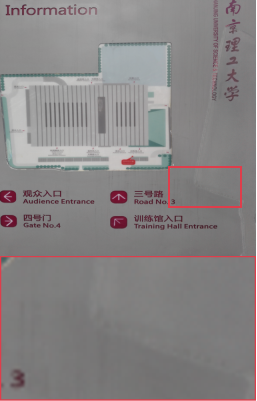}{(c) SGShadowNet~\cite{wan2022style}}\hspace{\gutter}%
  \imgcellcap{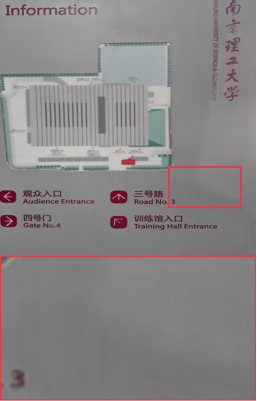}{\shortstack[c]{(d) Inpaint4-\\Shadow~\cite{li2023leveraging}}}\hspace{\gutter}%
  \imgcellcap{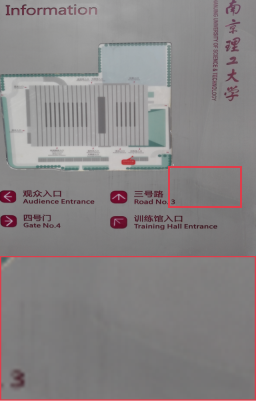}{(e) HomoFormer~\cite{xiao2024homoformer}}\hspace{\gutter}%
  \imgcellcap{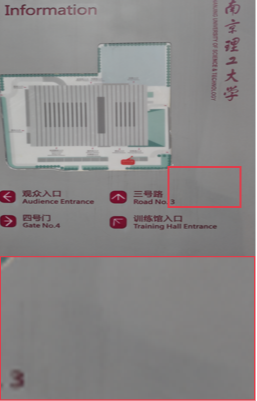}{\shortstack[c]{(f) StableShadow-\\Removal~\cite{xu_2025_CVPR}}}\hspace{\gutter}%
  \imgcellcap{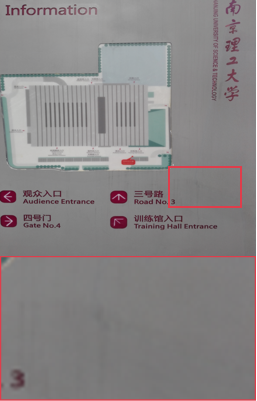}{(g) Ours}\hspace{\gutter}%
  \imgcellcap{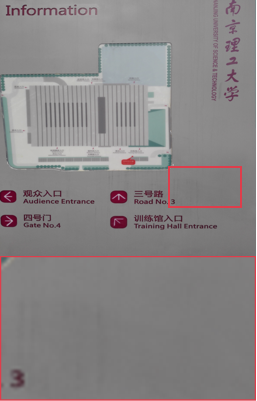}{(h) Ground Truth}%

  \caption{Visual comparisons with state-of-the-art methods on the ISTD+ dataset. (Best viewed zoomed in.)}
  \label{fig:exp_istd+}
\end{figure*}

\subsubsection{Evaluation on the ISTD+ Dataset} We further validate the effectiveness of our method on the ISTD+ dataset, with quantitative results presented in Tab.~\ref{tab:exp_istd+}. DeshadowMamba achieves consistent improvements over existing approaches across all evaluation metrics. It surpasses all competitors in terms of RMSE and PSNR in shadowed, non-shadowed, and overall regions, except for a negligible $0.002$ drop in SSIM on non-shadow areas compared to ShadowDiffusion~\cite{guo2023shadowdiffusion}.
In particular, in shadow regions, our method improves RMSE by $0.08$ and PSNR by $0.80$ dB over the second-best competitor StableShadowRemoval~\cite{xu_2025_CVPR}, highlighting its superior effectiveness in accurately restoring shadowed areas and maintaining overall image fidelity.

As shown in Fig.~\ref{fig:exp_istd+}, previous methods often leave visible artifacts and color inconsistencies near shadow boundaries, whereas DeshadowMamba produces smoother transitions and more faithful illumination restoration.
Notably, while some prior methods (\eg, HomoFormer~\cite{xiao2024homoformer} and StableShadowRemoval~\cite{xu_2025_CVPR}) exhibit considerable performance variations across datasets, DeshadowMamba consistently ranks first on both SRD and ISTD+, reflecting its strong robustness and adaptability to diverse data distributions.

\begin{table}[t]\centering
\setlength{\tabcolsep}{2pt} 
\caption{Quantitative evaluation with state-of-the-art methods on the SBU dataset.}
\label{supp_tab:quan_sbu}
    \begin{tabular}{l|c c}
        \toprule
        Method & NIQE$\downarrow$ & NIMA$\uparrow$\\
        \midrule
        BMNet~\cite{zhu2022bijective} & 4.00 & 4.40 \\
        ShadowFormer~\cite{guo2023shadowformer} & 3.97 & 4.45 \\
        Homoformer~\cite{xiao2024homoformer}& 3.94 & 4.46 \\
        StableShadowRemoval~\cite{xu_2025_CVPR} & 3.96 & 4.32 \\
        Ours & \textbf{3.89} & \textbf{4.51}\\
        \bottomrule
    \end{tabular}
\end{table}

\subsubsection{Evaluation on Real-World Shadow Removal}
\begin{figure*}[htbp]
    \captionsetup{skip=2pt}
    \centering
    \begin{minipage}{0.18\textwidth}
        \centering
        \includegraphics[width=\textwidth]{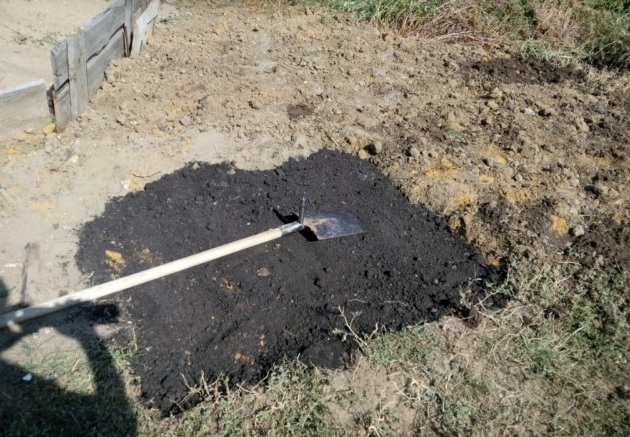}
    \end{minipage}
    \hspace{0.003\textwidth}
    \begin{minipage}{0.18\textwidth}
        \centering
        \includegraphics[width=\textwidth]{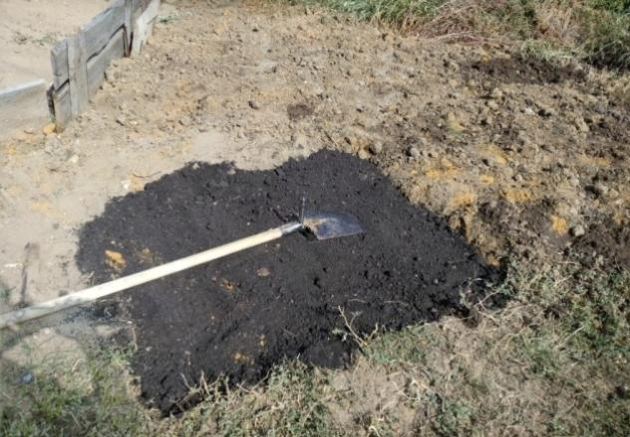}
    \end{minipage}
    \hspace{0.003\textwidth}
    \begin{minipage}{0.18\textwidth}
        \centering
        \includegraphics[width=\textwidth]{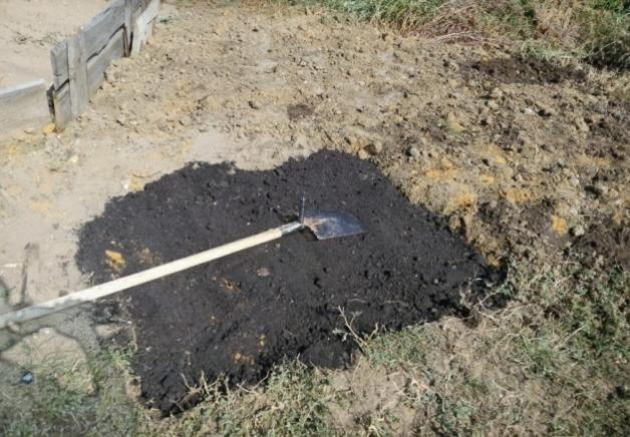}
    \end{minipage}
    \hspace{0.003\textwidth}
    \begin{minipage}{0.18\textwidth}
        \centering
        \includegraphics[width=\textwidth]{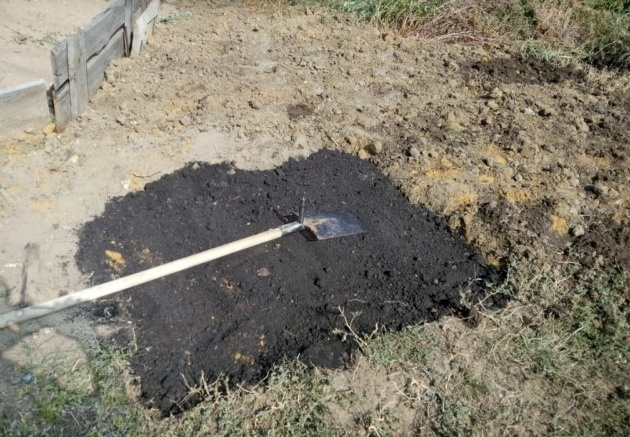}
    \end{minipage}
    \hspace{0.003\textwidth}
    \begin{minipage}{0.18\textwidth}
        \centering
        \includegraphics[width=\textwidth]{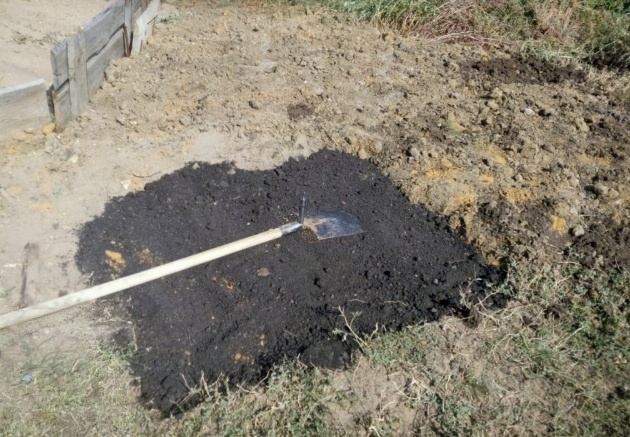}
    \end{minipage}
    
    \vspace{0.1cm}

    \begin{minipage}{0.18\textwidth}
        \centering
        \includegraphics[width=\textwidth]{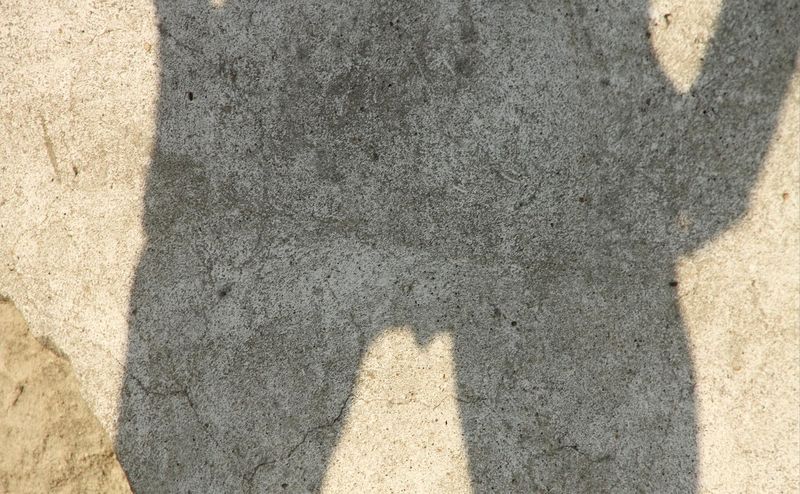}
        \caption*{\scriptsize (a) Shadow Image}
    \end{minipage}
    \hspace{0.003\textwidth}
    \begin{minipage}{0.18\textwidth}
        \centering
        \includegraphics[width=\textwidth]{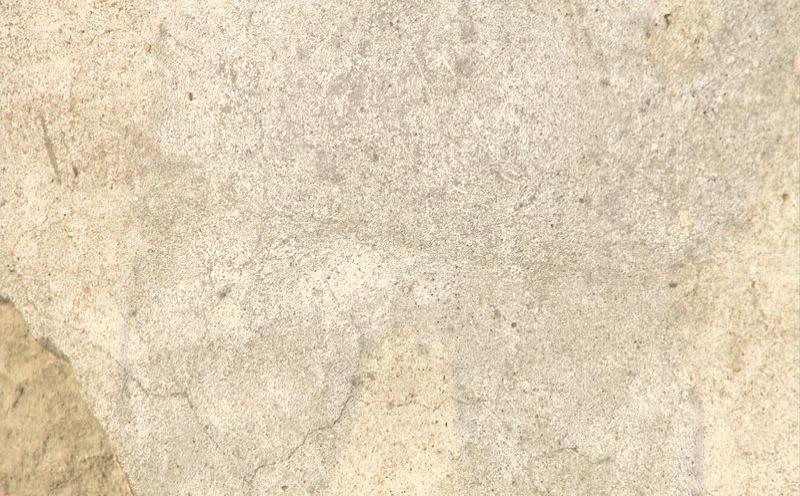}
        \caption*{\scriptsize (b) BMNet~\cite{zhu2022bijective}}
    \end{minipage}
    \hspace{0.003\textwidth}
    \begin{minipage}{0.18\textwidth}
        \centering
        \includegraphics[width=\textwidth]{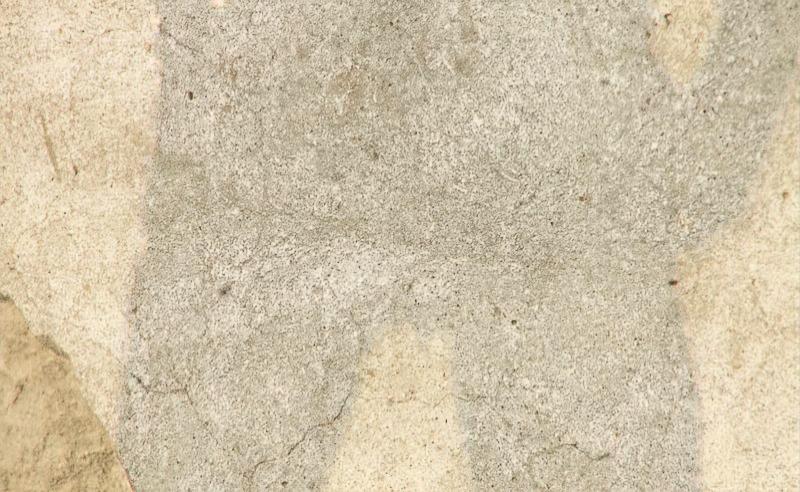}
        \caption*{\scriptsize (c) SGShadowNet~\cite{wan2022style}}
    \end{minipage}
    \hspace{0.003\textwidth}
    \begin{minipage}{0.18\textwidth}
        \centering
        \includegraphics[width=\textwidth]{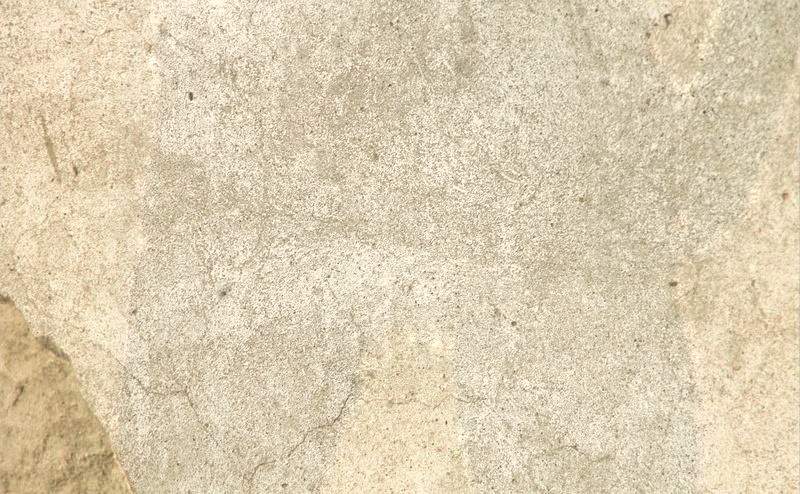}
        \caption*{\scriptsize (d) HomoFormer~\cite{xiao2024homoformer}}
    \end{minipage}
    \hspace{0.003\textwidth}
    \begin{minipage}{0.18\textwidth}
        \centering
        \includegraphics[width=\textwidth]{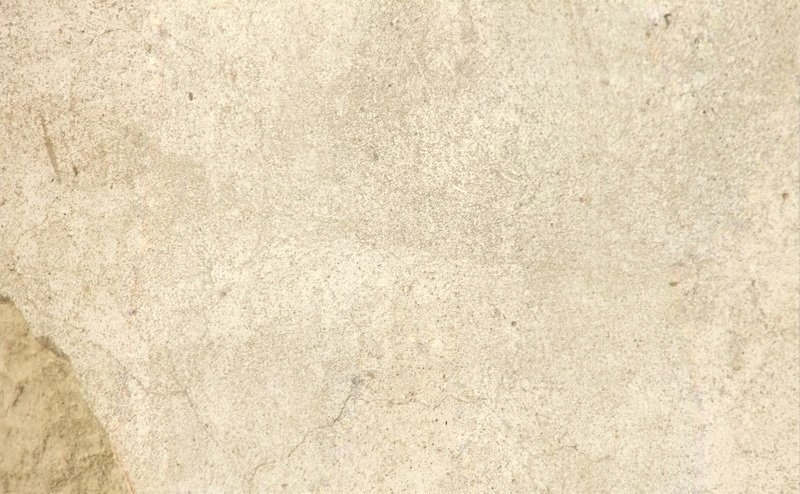}
        \caption*{\scriptsize (f) Ours}
    \end{minipage}    
    \caption{Visual comparisons with state-of-the-art methods for real-world shadow removal on the SBU dataset.}
    \label{fig:exp_sbu}
\end{figure*}

To further assess the generalization ability of our method in real-world conditions, we evaluate DeshadowMamba on the SBU dataset~\cite{vicente2016large}, which contains real-world shadow images without ground truth images. All methods are evaluated using models trained on ISTD+~\cite{le2019shadow}.
As shown in Fig.~\ref{fig:exp_sbu}, existing methods often fail to completely remove shadows or leave noticeable residuals along shadow boundaries, whereas DeshadowMamba effectively eliminates both cast and self-shadows, producing cleaner and more visually consistent results across various real-world scenes.

To complement the qualitative study, we also report no-reference perceptual quality metrics, NIQE and NIMA~\cite{talebi2018nima}, on the SBU dataset. NIQE is computed on the luminance channel using default settings, and NIMA is obtained from the official aesthetic predictor. As summarized in Tab.~\ref{supp_tab:quan_sbu}, our method achieves the best NIQE and NIMA among all compared approaches, indicating fewer perceptual distortions and higher aesthetic consistency in complex, ground-truth–absent real-world scenes.

\begin{table}[t]
    \centering
    \setlength{\tabcolsep}{2pt} 
    \caption{Ablation of CrossGate modulation on SRD. All variants are trained without ColorShift regularization.}
    \label{tab:ablation_cross}
    \begin{tabular}{l|cc|cc}
        \toprule
        \multirow{2}{*}{Setting} & \multicolumn{2}{c|}{Shadow} & \multicolumn{2}{c}{All Image} \\
        \cmidrule(r){2-3} \cmidrule(r){4-5}
        & RMSE$\downarrow$ & PSNR$\uparrow$ & RMSE$\downarrow$ & PSNR$\uparrow$ \\
        \midrule
        Baseline & 4.24 & 38.50 & 3.15 & 35.42\\
        w/ $\bm{G}_h$ \& w/o $\bm{G}_v$ & 4.22 & 38.65 & 3.13 & 35.54 \\
        w/ $\bm{G}_v$ \& w/o $\bm{G}_h$ & 4.21 & 38.68 & 3.12 & 35.56 \\ 
        w/o offset predictor & 4.28 & 38.50 & 3.13 & 35.45\\
        \midrule
        Ours & \textbf{4.17} & \textbf{38.75} & \textbf{3.09} & \textbf{35.63}\\
        \bottomrule
    \end{tabular}
\end{table}
\subsection{Ablation Study}

\begin{figure}[tbp]
    \centering
    \includegraphics[width=0.47\textwidth]{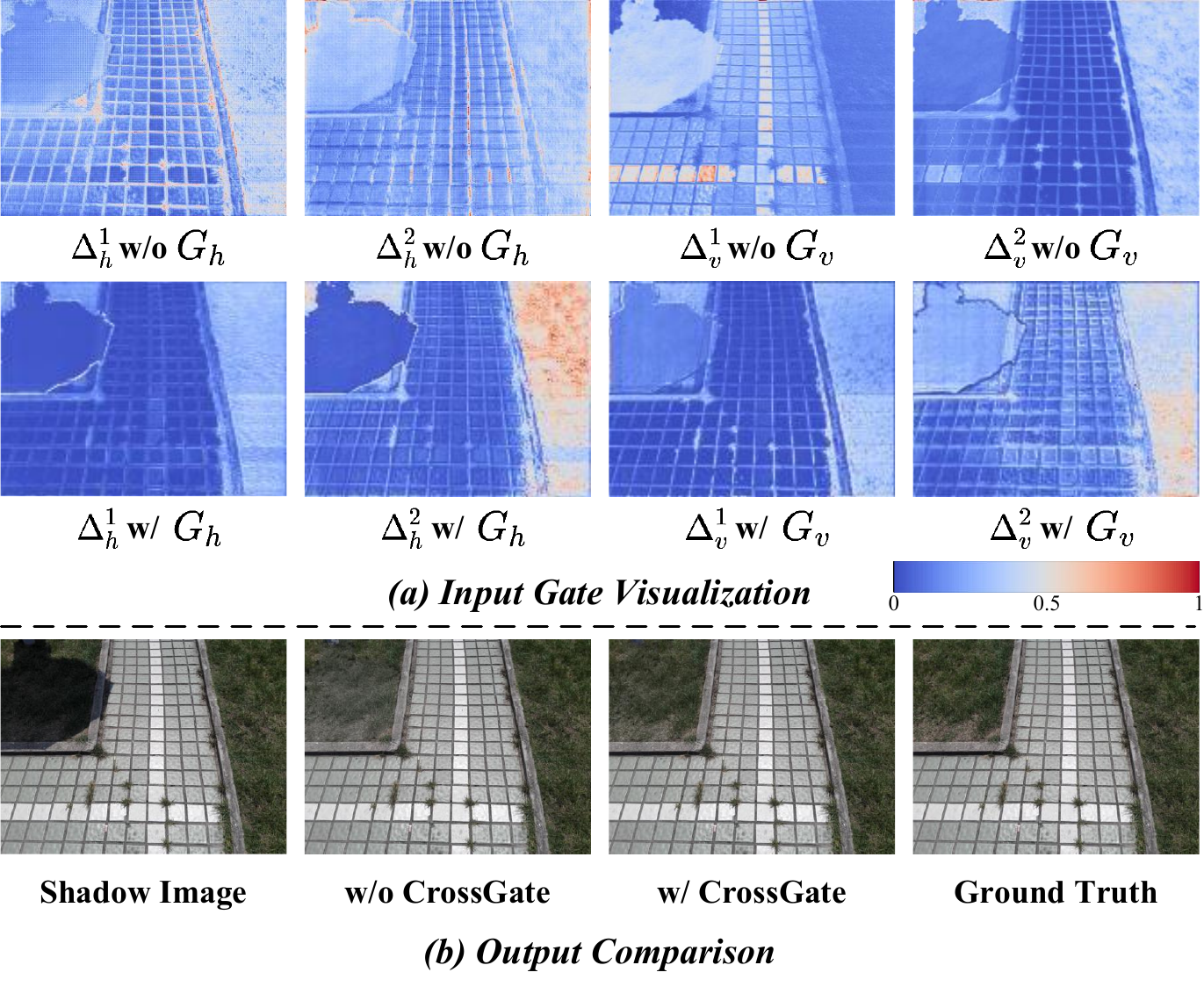}
    \caption{Visual effects of CrossGate modulation on input gates (final Mamba block) and deshadowing results.} \vspace{-5mm}
    \label{fig:ablation_cross}
\end{figure}

\subsubsection{Effectiveness of CrossGate Modulation} To validate the effectiveness of our CrossGate modulation, we conduct an ablation study with several variants on SRD, summarized in Tab.~\ref{tab:ablation_cross}. We begin with a baseline model built upon the Visual State Space Module (VSSM)~\cite{liu2024vmamba}. Even when directly applied to shadow removal, this baseline already outperforms the competitive transformer-based method HomoFormer~\cite{xiao2024homoformer}, confirming that Mamba is a highly suitable backbone for this task. We then enhance this baseline by independently incorporating horizontal and vertical gate modulations, both of which lead to performance improvements. Next, we examine the necessity of the deformable sampling via a variant without the offset predictor. In this case, the query and unwarped key maps $\bm{Q}$ and $\bm{K}$ directly compute directional point-wise similarity. The results show that such fixed-pattern similarity computation not only provides no benefits but also degrades performance in shadow regions, mainly due to the possible lack of semantically relevant areas along the same row or column. This finding further justifies our design choice of introducing the offset predictor. With the full CrossGate modulation applied, the model captures richer non-local dependencies and ultimately achieves the best performance, demonstrating the value of incorporating input gate modulation into Mamba.

Fig.~\ref{fig:ablation_cross} visualizes the impact of CrossGate modulation on Mamba's input gates. We show gate activations from the final Mamba block along four scanning paths: top-left to bottom-right ($\bm{\Delta}_h^1$), bottom-right to top-left ($\bm{\Delta}_h^2$), and their 90-degree rotated counterparts ($\bm{\Delta}_v^1$, $\bm{\Delta}_v^2$), both with and without the modulation signals $\bm{G}_h$ and $\bm{G}_v$. The visualization clearly indicates that CrossGate alters the activation distribution during scanning. Without modulation, the network shows limited sensitivity to shadow regions and contextually relevant areas. For instance, in $\bm{\Delta}_h^1$ without $\bm{G}_h$, the right-side lawn and central tiles receive similarly high responses, weakening the model’s ability to form meaningful semantic associations. In contrast, with CrossGate enabled, the modulation signals highlight non-shadow regions that are semantically correlated with shadow queries, guiding the model to rely more on the input signal rather than the hidden state during scanning and thus providing richer cues for subsequent shadow restoration. These observations verify the role of CrossGate in improving semantic correspondence during shadow removal.

\begin{table}[t]
    \centering
    \setlength{\tabcolsep}{2pt} 
    \caption{Ablation of ColorShift regularization on SRD. All variants are equipped with CrossGate modulation.}
    \label{tab:ablation_cs}
    \begin{tabular}{l|cc|cc}
        \toprule
        \multirow{2}{*}{Setting} & \multicolumn{2}{c|}{Shadow} & \multicolumn{2}{c}{All Image} \\
        \cmidrule(r){2-3} \cmidrule(r){4-5}
        & RMSE$\downarrow$ & PSNR$\uparrow$ & RMSE$\downarrow$ & PSNR$\uparrow$ \\
        \midrule
        w/o CS & 4.17 & 38.75 & 3.09 & 35.63 \\        
        CR~\cite{wu2021contrastive} & 4.15 & 38.99 & 3.14 & 35.67 \\
        \midrule
        \multicolumn{5}{c}{Negative Sample Generation Strategy} \\
        \midrule
        Random Colors & 4.20 & 38.75 & 3.13 & 35.51 \\
        Exposure Perturbation& 4.29 & 38.43 & 3.19 & 35.22  \\    
        \midrule
        \multicolumn{5}{c}{Negative Sample Weighting Scheme} \\
        \midrule
        Uniform& 4.23 & 38.64 & 3.13 & 35.45 \\
        Direct-Normalized & 4.23 & 38.65 & 3.13 & 35.46  \\ 
        \midrule
        Ours & \textbf{4.09} & \textbf{39.17} & \textbf{3.04} & \textbf{35.94}\\
        \bottomrule
    \end{tabular}
\end{table}

\subsubsection{Effectiveness of ColorShift Regularization} We further assess the impact of our ColorShift (CS) Regularization in Tab.~\ref{tab:ablation_cs}. We begin by applying a naive contrastive regularization (CR) strategy~\cite{wu2021contrastive}, which samples shadow patches from different images in the training set as negative samples and assigns them uniform weights. Surprisingly, this approach underperforms even compared to using no regularization. This is likely due to the fact that shadow removal targets localized degradations, which fundamentally differ from the global patterns seen in tasks like dehazing. As a result, many contrastive pairs exhibit significant degradation misalignment, rendering this CR strategy ineffective for shadow removal.

Next, we replace the color clustering module in CS with $K$ randomly generated colors while keeping all other settings fixed. This results in a performance drop, suggesting that spurious color cues can misguide the model. It also supports the notion that color contamination in shadowed regions mainly stems from entanglement with irrelevant background colors. In addition, we evaluate a negative sampling method based on under/over-exposure~\cite{liang2022semantically}, which leads to an even larger degradation. In contrast, our CS formulation, by considering degradation-aware color semantics and constructing task-aligned negative samples, achieves better results.

Finally, we explore alternative weighting schemes in CS. Assigning uniform weights to all negatives leads to clear performance degradation, largely due to varying numbers of negative samples across instances and the resulting training instability. Similarly, directly normalizing the difficulty metric $\{R_i\}$ reduces performance, as it causes the model to neglect harder negatives. By comparison, our reciprocal-normalized weighting strategy maintains balanced attention across shifted color negatives, yielding consistently superior results. 

\begin{table*} 
	\centering
    \setlength{\tabcolsep}{2pt} 
        \caption{Generalizability of ColorShift regularization across state-of-the-art methods on ISTD+.}
        \label{supp_tab:gen_coloshift}
	\begin{tabular}{c|cc|cc|cc}
		\toprule
		\multirow{2}{*}{Method} & \multicolumn{2}{c|}{Shadow} & \multicolumn{2}{c|}{Non-Shadow} & \multicolumn{2}{c}{All Image} \\ 
		 \cmidrule(lr){2-3} \cmidrule(lr){4-5} \cmidrule(lr){6-7}
         & RMSE$\downarrow$ & PSNR$\uparrow$
	& RMSE$\downarrow$ & PSNR$\uparrow$ 
	& RMSE$\downarrow$ & PSNR$\uparrow$ \\ 
    \hline
		SGShadowNet~\cite{wan2022style}             & 6.46         & 36.91  & 2.95        & 35.47            & 3.45     & 32.46       \\
		\rowcolor[HTML]{f2f2f2}
		SGShadowNet+CS          & 6.00(\textcolor{red}{0.46$\downarrow$})  & 37.66 (\textcolor{red}{0.75$\uparrow$})          &    2.47 (\textcolor{red}{0.48$\downarrow$})         &   37.55 (\textcolor{red}{2.11$\uparrow$})                  & 3.01 (\textcolor{red}{0.44$\downarrow$})         &  34.00 (\textcolor{red}{1.54$\uparrow$})               \\ 
        \hline
		ShadowFormer~\cite{guo2023shadowformer} &      5.34           &  39.54          &     2.34        &   38.72                   &    2.81      &   35.44              \\
		\rowcolor[HTML]{f2f2f2}
		ShadowFormer+CS&    5.16 (\textcolor{red}{0.18$\downarrow$})                &  39.78 (\textcolor{red}{0.24$\uparrow$})         &  2.30 (\textcolor{red}{0.04$\downarrow$})          &  38.78 (\textcolor{red}{0.06$\uparrow$})                 &   2.75 (\textcolor{red}{0.06$\downarrow$})      &  35.60 (\textcolor{red}{0.16$\uparrow$})               \\ \hline
		HomoFormer~\cite{xiao2024homoformer}&      4.92               &  39.51          &    2.27         &     38.65                 &    2.68      &    35.32            \\
		\rowcolor[HTML]{f2f2f2}
		HomoFormer+CS&       4.58 (\textcolor{red}{0.34$\downarrow$})             &  40.68 (\textcolor{red}{1.17$\uparrow$})      &     2.20 (\textcolor{red}{0.07$\downarrow$})       &   39.10 (\textcolor{red}{0.45$\uparrow$})               &   2.58 (\textcolor{red}{0.10$\downarrow$})      &   36.08 (\textcolor{red}{0.76$\uparrow$})        \\ \hline
	\end{tabular}
\end{table*}

\begin{figure}[htbp]
    \captionsetup{skip=2pt}
    \centering
    \begin{minipage}{0.11\textwidth}
        \centering
        \includegraphics[width=\textwidth]{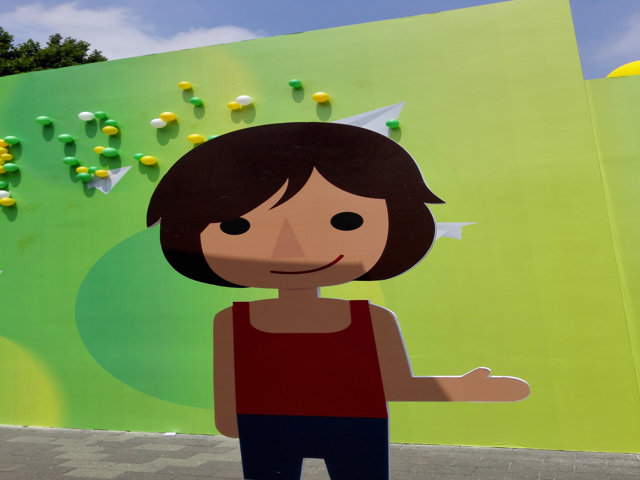}
        \caption*{\scriptsize Shadow Image}
    \end{minipage}
    \hspace{0.002\textwidth}
    \begin{minipage}{0.11\textwidth}
        \centering
        \includegraphics[width=\textwidth]{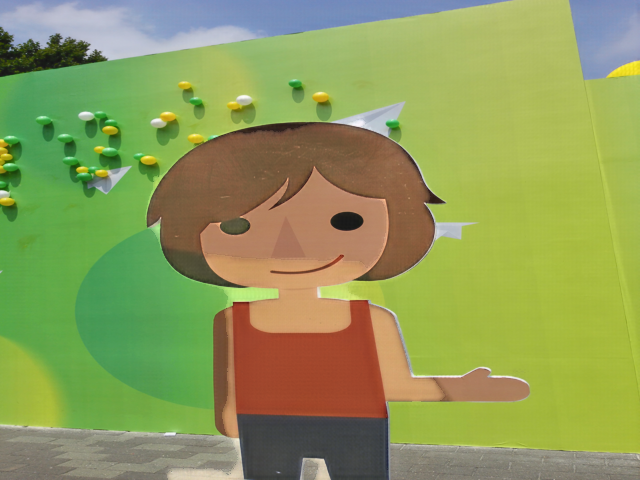}
        \caption*{\scriptsize SGShadowNet~\cite{wan2022style}}
    \end{minipage}
    \hspace{0.002\textwidth}
    \begin{minipage}{0.11\textwidth}
        \centering
        \includegraphics[width=\textwidth]{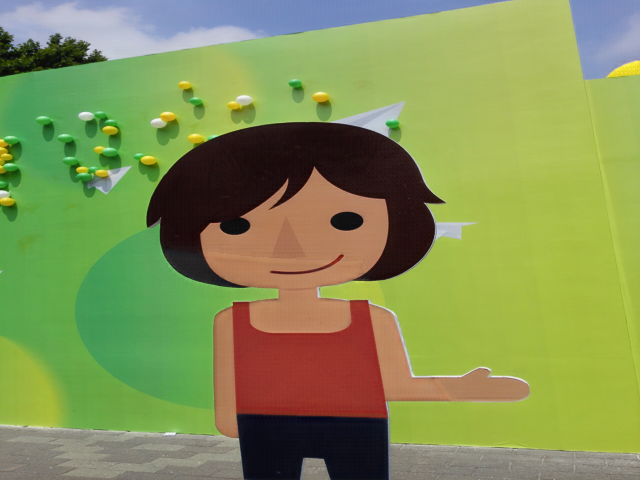}
        \caption*{\scriptsize SGShadowNet+CS}
    \end{minipage}
    \hspace{0.002\textwidth}
    \begin{minipage}{0.11\textwidth}
        \centering
        \includegraphics[width=\textwidth]{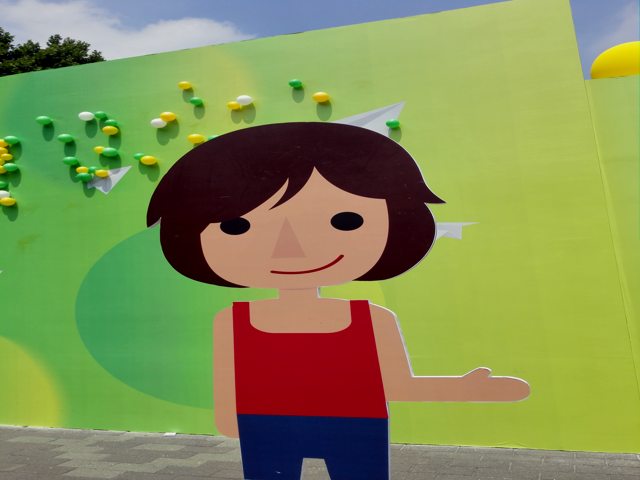}
        \caption*{\scriptsize Ground Truth}
    \end{minipage}
    \vspace{0.05cm}
    \begin{minipage}{0.11\textwidth}
        \centering
        \includegraphics[width=\textwidth]{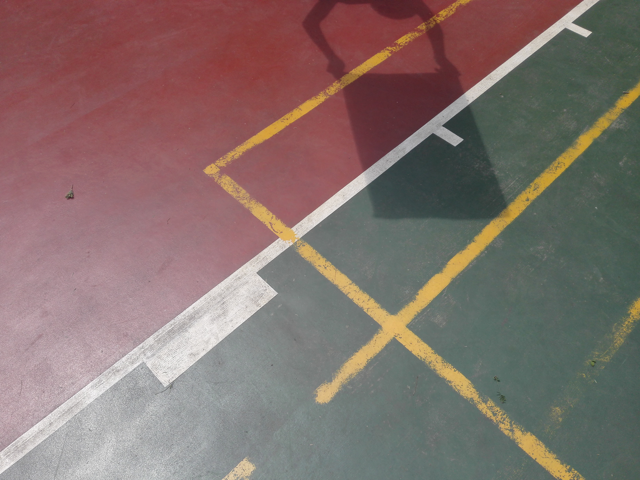}
        \caption*{\scriptsize Shadow Image}
    \end{minipage}
    \hspace{0.002\textwidth}
    \begin{minipage}{0.11\textwidth}
        \centering
        \includegraphics[width=\textwidth]{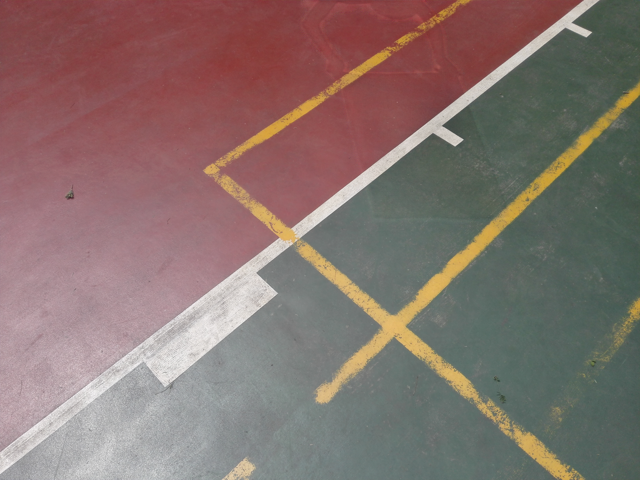}
        \caption*{\scriptsize ShadowFormer~\cite{guo2023shadowformer}}
    \end{minipage}
    \hspace{0.002\textwidth}
    \begin{minipage}{0.11\textwidth}
        \centering
        \includegraphics[width=\textwidth]{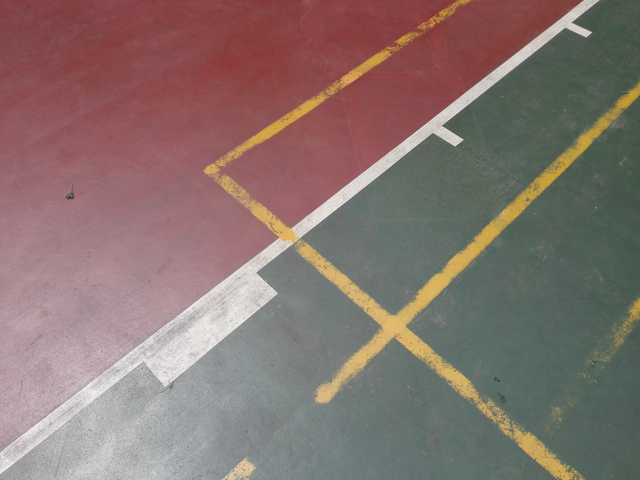}
        \caption*{\scriptsize ShadowFormer+CS}
    \end{minipage}
    \hspace{0.002\textwidth}
    \begin{minipage}{0.11\textwidth}
        \centering
        \includegraphics[width=\textwidth]{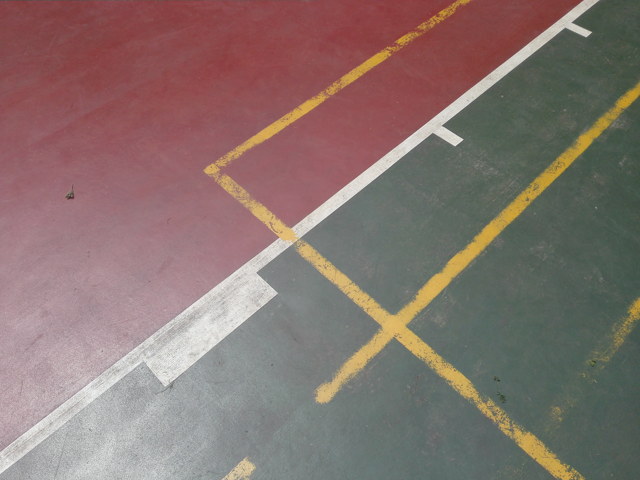}
        \caption*{\scriptsize Ground Truth}
    \end{minipage}
    \vspace{0.05cm}
    \begin{minipage}{0.11\textwidth}
        \centering
        \includegraphics[width=\textwidth]{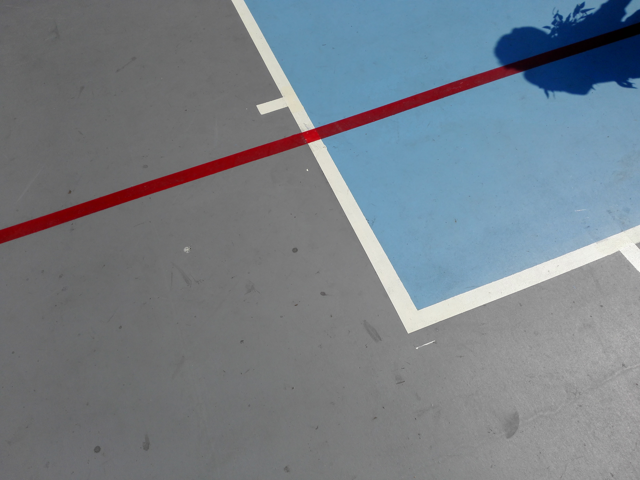}
        \caption*{\scriptsize Shadow Image}
    \end{minipage}
    \hspace{0.002\textwidth}
    \begin{minipage}{0.11\textwidth}
        \centering
        \includegraphics[width=\textwidth]{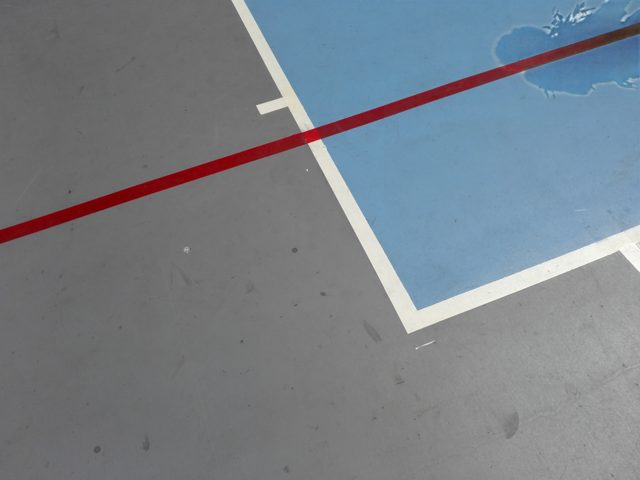}
        \caption*{\scriptsize HomoFormer~\cite{xiao2024homoformer}}
    \end{minipage}
    \hspace{0.002\textwidth}
    \begin{minipage}{0.11\textwidth}
        \centering
        \includegraphics[width=\textwidth]{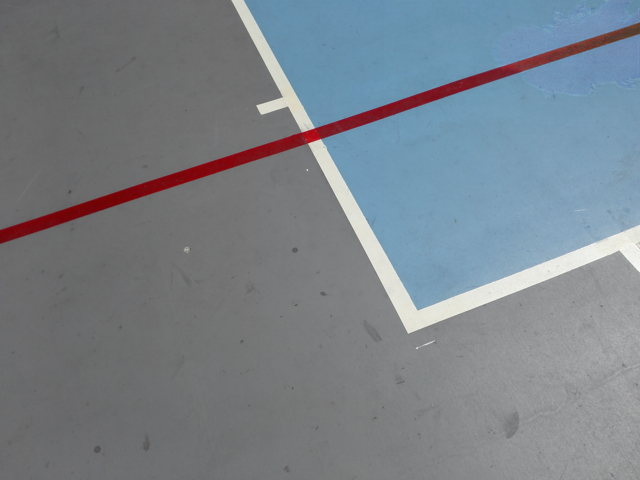}
        \caption*{\scriptsize HomoFormer+CS}
    \end{minipage}
    \hspace{0.002\textwidth}
    \begin{minipage}{0.11\textwidth}
        \centering
        \includegraphics[width=\textwidth]{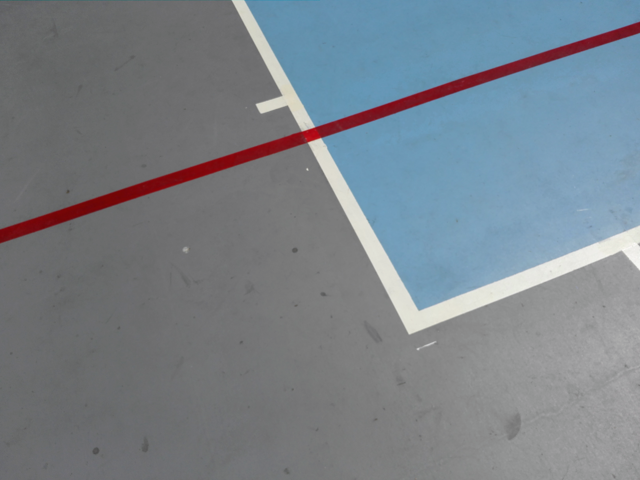}
        \caption*{\scriptsize Ground Truth}
    \end{minipage}
    \caption{Visual comparisons of ColorShift regularization applied to state-of-the-art methods on ISTD+. (Best viewed zoomed in.)}
    \label{supfig:cstoothers}
\end{figure}

\subsection{Generalizability Analysis}
To evaluate the generalizability of the proposed ColorShift Regularization, we integrate it into three representative state-of-the-art methods: SGShadowNet~\cite{wan2022style}, ShadowFormer~\cite{guo2023shadowformer}, and HomoFormer~\cite{xiao2024homoformer}, all of which rely on global context modeling and thus suffer from color contamination introduced by whole-image statistics.
The quantitative results on the ISTD+ dataset, presented in Tab.~\ref{supp_tab:gen_coloshift}, show that our CS strategy consistently improves performance across all three methods, with most evaluation metrics exhibiting notable gains while introducing no additional parameters. The smallest improvement occurs with ShadowFormer. Although ShadowFormer employs channel attention to partially capture global dependencies, its Shadow-Interaction module is constrained by the shift-window mechanism, which limits its ability to exploit long-range context. Consequently, it is less affected by global color shifts, resulting in relatively modest improvements when CS is applied. In contrast, SGShadowNet and HomoFormer benefit more from our CS regularization, further demonstrating its strong generalization ability in models that leverage extensive contextual information.

Fig.~\ref{supfig:cstoothers} provides corresponding visual comparisons. As shown, CS effectively mitigates chromatic bias in models that emphasize long-range context aggregation, reinforcing its broad applicability across different network architectures. More detailed analyses are provided in the supplementary material.

\section{Limitation}
\begin{figure}
    \captionsetup{skip=2pt}
    \centering
    \begin{minipage}{0.15\textwidth}
        \centering
        \includegraphics[width=\textwidth]{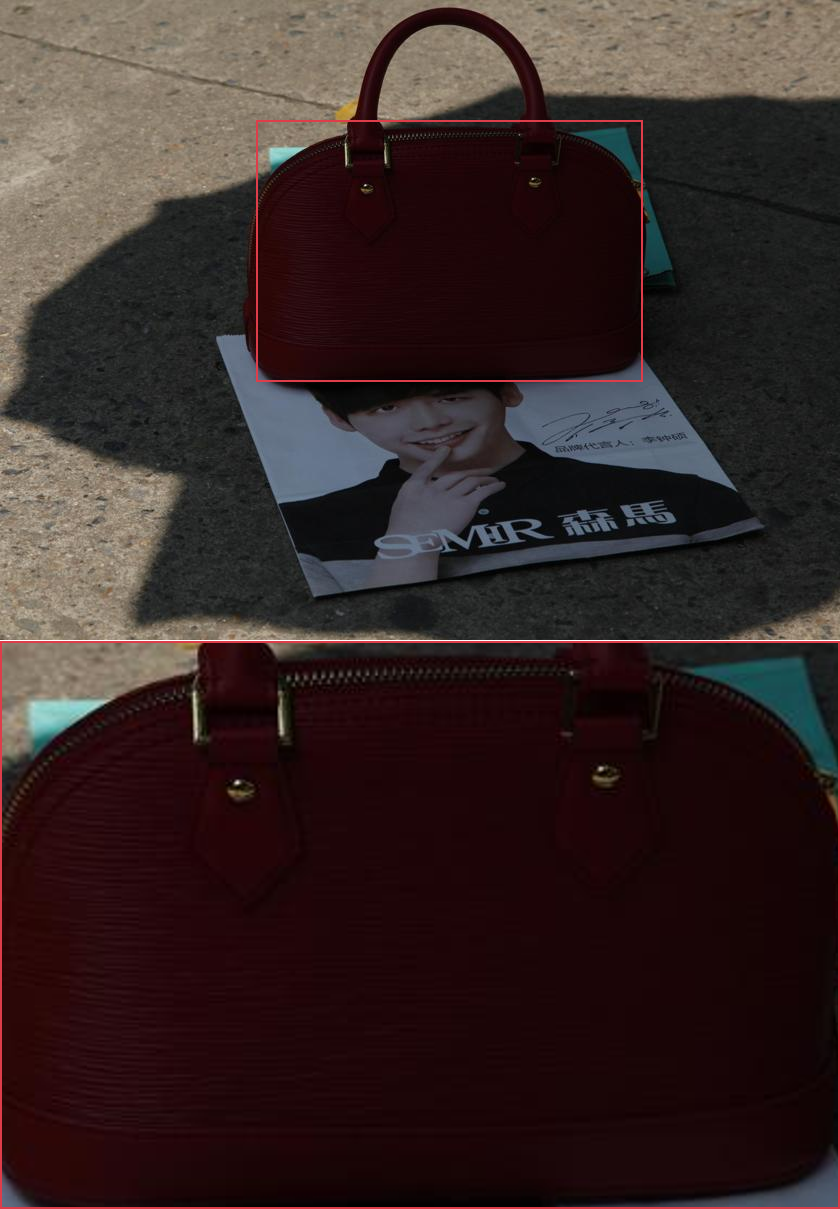}
        \caption*{\scriptsize (a) Shadow Image}
    \end{minipage}
    \hspace{0.003\textwidth}      
    \begin{minipage}{0.15\textwidth}
        \centering
        \includegraphics[width=\textwidth]{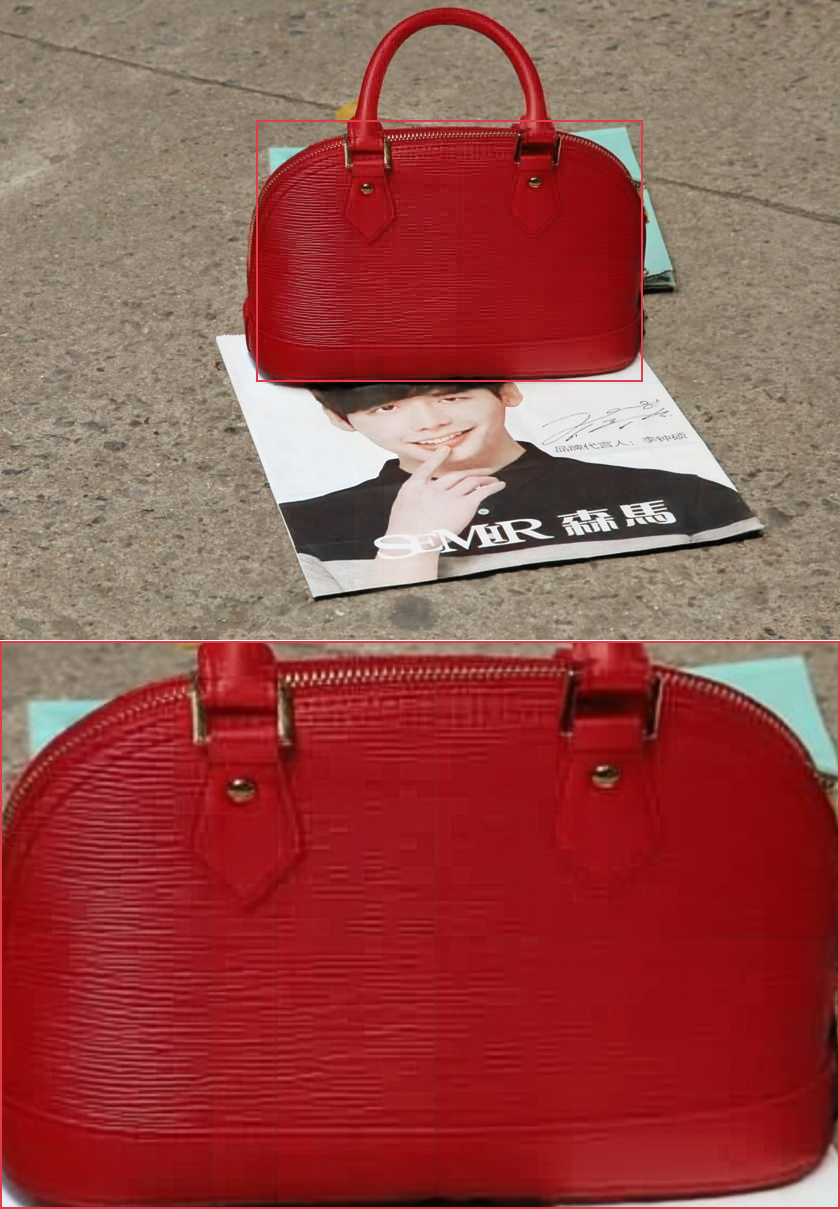}
        \caption*{\scriptsize (d) DeshadowMamba}
    \end{minipage}
    \hspace{0.003\textwidth}
    \begin{minipage}{0.15\textwidth}
        \centering
        \includegraphics[width=\textwidth]{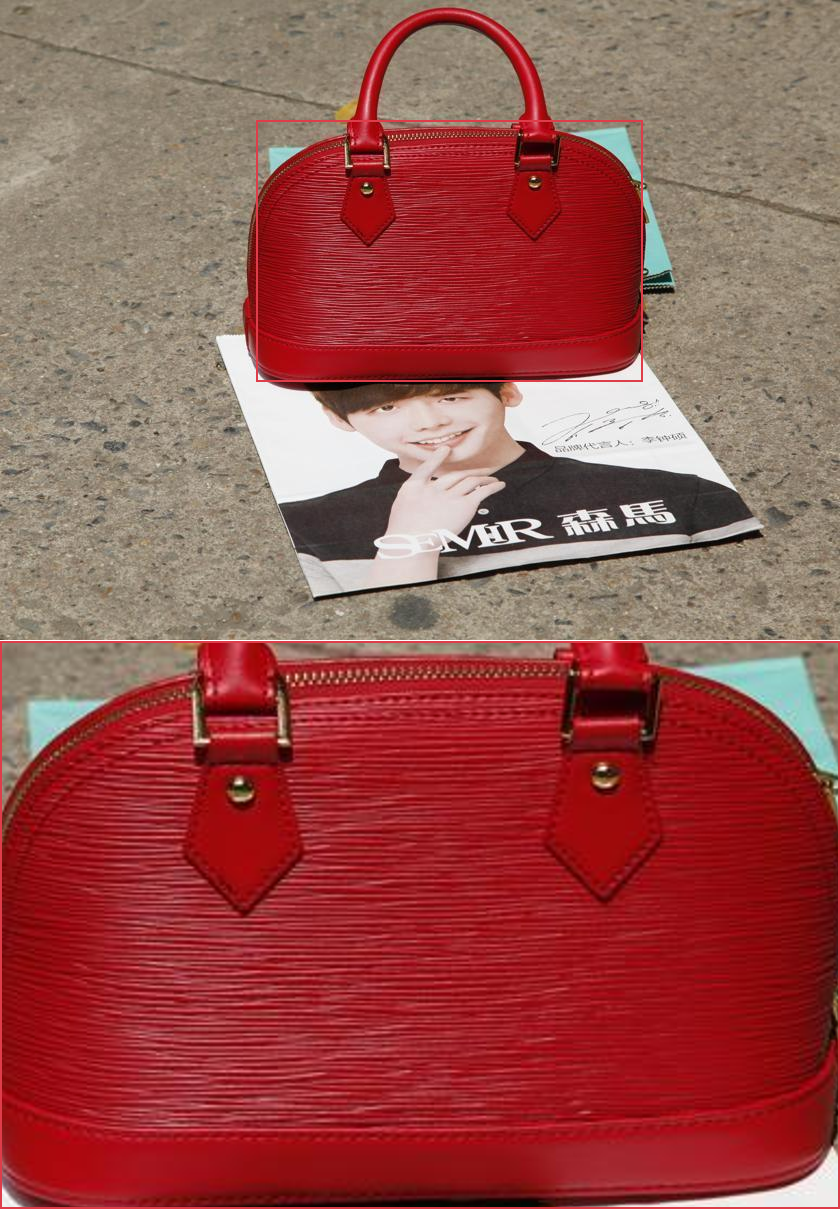}
        \caption*{\scriptsize (e) Ground Truth}
    \end{minipage}    
    \caption{Visual example of DeshadowMamba's limitation.}   
    \label{supfig:limitation}
\end{figure}
Despite the remarkable performance and robustness of DeshadowMamba, it still encounters a prevalent challenge in the shadow removal, which is the accurate restoration of locally occluded regions where no reliable shadow-free cues are available. When the shadowed area lacks meaningful correlations with non-shadow regions, our model cannot obtain sufficient contextual information for faithful reconstruction. As shown in Fig.~\ref{supfig:limitation}, DeshadowMamba struggles to recover fine textures on the red leather bag, primarily due to the absence of long-range or spatially adjacent informative guidance. This limitation suggests a promising direction for future work, such as leveraging generative priors or cross-image correspondence to supplement missing cues.

\section{Conclusion}
In this paper, we revisit the shadow removal problem through the lens of sequence modeling and present DeshadowMamba, a framework built upon Mamba’s state space modeling for efficient and structure-aware shadow removal. CrossGate serves as an input gate modulation module that captures spatial similarity across non-local regions, enabling context-aware integration of informative non-shadow cues. ColorShift regularization introduces a contrastive learning strategy guided by global color statistics, effectively mitigating color contamination and improving chromatic consistency. Extensive experiments on multiple benchmark datasets demonstrate the superiority of our approach, establishing new state-of-the-art results in both qualitative and quantitative evaluations.

\bibliographystyle{IEEEtran}
\bibliography{sample-base}

\newpage

\vfill

\end{document}